%% file: arxiv.tex
\documentclass[10pt,twocolumn,letterpaper]{article}

\usepackage[pagenumbers]{cvpr} % To force page numbers, e.g. for an arXiv version


\usepackage{graphicx}
\usepackage{subcaption}
\graphicspath{{./figures/}}
\usepackage{amsmath}
\usepackage{amssymb}
\usepackage{booktabs}
\usepackage{multirow}
\usepackage{booktabs}
\usepackage{amsfonts}
\usepackage{float}
\usepackage{makecell}
\usepackage{flushend, cuted}
\usepackage{xcolor}
\usepackage{colortbl}
\usepackage{array}

\renewcommand{\figurename}{Fig.}

\definecolor{cvprblue}{rgb}{0.21,0.49,0.74}
\usepackage[pagebackref,breaklinks,colorlinks,allcolors=cvprblue]{hyperref}

\def\paperID{642} % *** Enter the Paper ID here
\def\confName{CVPR}
\def\confYear{2025}

\title{Toward Generalized Image Quality Assessment: \\ Relaxing the Perfect Reference Quality Assumption}

\author{
	Du Chen$^{1,3,}$\thanks{Equal contribution.}, \
	Tianhe Wu$^{2,3,}$\footnotemark[1], \
	Kede Ma$^{2,}$\thanks{Corresponding authors.}, and 
	Lei Zhang$^{1,3,}$\footnotemark[2]\\
	$^1$The Hong Kong Polytechnic University \ 
	$^2$City University of Hong Kong \  $^3$OPPO Research Institute\\
	{\tt\small csdud.chen@connet.polyu.hk}, {\tt\small \{tianhewu, kede.ma\}@cityu.edu.hk}, {\tt\small cslzhang@comp.polyu.edu.hk}\\
}
\begin{document}
\maketitle

\begin{strip}
	\centering
	\vspace{-60pt}
	\includegraphics[width=\textwidth]{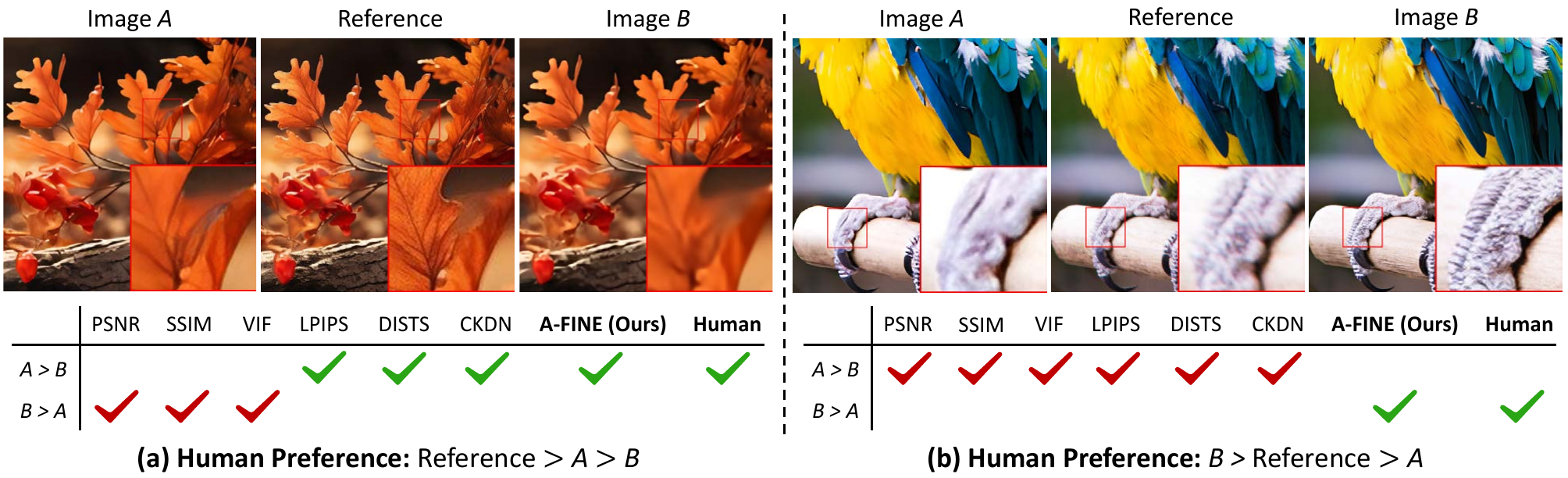}
	\vspace{-18pt}
	\captionof{figure}{
		\textit{With the reference image in the middle, which image, A or B, has better perceived visual quality}? The proposed A-FINE generalizes and outperforms standard FR-IQA models under both perfect and imperfect reference conditions. Zoom in for better visibility.}
	\label{fig:teaser}
\end{strip}

\begin{abstract}
Full-reference image quality assessment (FR-IQA) generally assumes that reference images are of perfect quality. However, this assumption is flawed due to the sensor and optical limitations of modern imaging systems. Moreover, recent generative enhancement methods are capable of producing images of higher quality than their original. All of these challenge the effectiveness and applicability of current FR-IQA models. To relax the assumption of perfect reference image  quality, we build a large-scale IQA database, namely DiffIQA, containing approximately $180,000$ images generated by a diffusion-based image enhancer with adjustable hyper-parameters. Each image is annotated by human subjects as either worse, similar, or better quality compared to its reference. Building on this, we present a generalized FR-IQA model, namely \textbf{A}daptive \textbf{FI}delity-\textbf{N}aturalness \textbf{E}valuator (A-FINE), to accurately assess and adaptively combine the fidelity and naturalness of a test image. A-FINE aligns well with standard FR-IQA when the reference image is much more natural than the test image. We demonstrate by extensive experiments that A-FINE surpasses standard FR-IQA models on well-established IQA datasets and our newly created DiffIQA. To further validate A-FINE, we additionally construct a super-resolution IQA benchmark (SRIQA-Bench), encompassing test images derived from ten state-of-the-art SR methods with reliable human quality annotations. Tests on SRIQA-Bench re-affirm the advantages of A-FINE. The code and dataset are available at \url{https://tianhewu.github.io/A-FINE-page.github.io/}.
\end{abstract}

\section{Introduction}
\label{sec:intro}
Image Quality Assessment (IQA) plays an indispensable role in the digital image lifecycle, from acquisition, transmission, and reproduction, to storage~\cite{wang2009mean}. The objective of IQA is to develop computational models that mimic the Human Visual System (HVS) in perceiving image quality~\cite{wang2004image}, which can be broadly classified into two categories based on the availability of reference images: Full-Reference IQA (FR-IQA)~\cite{wang2004image, wang2003multiscale, zhang2011fsim, zhang2018unreasonable, ding2020image, ding2021locally} and No-Reference IQA (NR-IQA)~\cite{mittal2012making, yang2022maniqa, zhang2021uncertainty, zhang2023blind, wang2011reduced, you2025teaching}. FR-IQA evaluates a test image by comparing it to a reference image, which is assumed to be of perfect quality, while NR-IQA assesses the test image quality without needing the reference.  

Over the past two decades, FR-IQA has experienced significant progress, with the paradigm shifted from measuring error visibility~\cite{daly1992visible} to assessing structural similarity~\cite{wang2004image}, and more recently, to unifying structural and textural similarity~\cite{ding2020image}. This evolution has led to the development of several representative methods, including SSIM~\cite{wang2004image}, FSIM~\cite{zhang2011fsim}, LPIPS~\cite{zhang2018unreasonable}, and DISTS~\cite{ding2020image}. These FR-IQA models have been rapidly adopted as standard evaluation criteria, alongside the traditional peak signal-to-noise ratio (PSNR)~\cite{wang2009mean}, for measuring the progress in various image processing tasks. Moreover, there is a growing trend of employing these models as loss functions for perceptual optimization of image restoration algorithms based on deep neural networks (DNNs)~\cite{ding2021comparison, ledig2017photo, zhang2019ranksrgan, liang2022details, chen2024ssl, sun2024pixel}.

\begin{figure}[t]
  \centering
  \begin{subfigure}[b]{0.23\textwidth}
    \includegraphics[width=\textwidth]{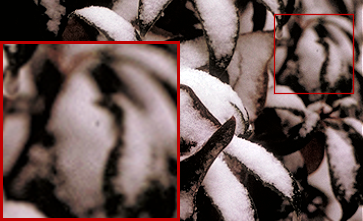}
    \caption{Reference image}
  \end{subfigure}
  % \hfill 
  \begin{subfigure}[b]{0.23\textwidth}
    \includegraphics[width=\textwidth]{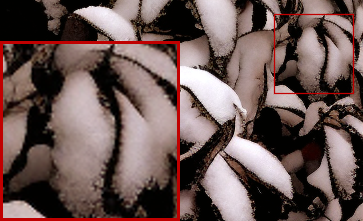}
    \caption{Enhanced image}
  \end{subfigure}
  \vspace{-2mm}
  	\caption{
 \textbf{(a)} Reference image from CSIQ \cite{larson2010most} and \textbf{(b)} its corresponding enhanced image by a recent generation-based image enhancer, SeeSR~\cite{wu2024seesr}.
 }
	\label{fig:phenomenon}
 \vspace{-3mm}
\end{figure}

Most FR-IQA models operate under the assumption that the reference image is of perfect quality. However, this assumption is problematic as digital imaging systems face practical hardware and software limitations, making it extremely difficult (if not impossible) to capture perfect-quality images. This is particularly true for natural scenes that exhibit great spatiotemporal complexity, high dynamic range, and wide color spectrum. As a result, many reference images in existing IQA datasets are of subpar quality. Moreover, the image quality could be rescued and even improved using modern generative image enhancement techniques~\cite{wang2024exploiting,yang2023pixel,sun2023improving,yu2024scaling,wu2024one} (see \figurename~\ref{fig:phenomenon} for a visual example).

\begin{figure}[t]
	\centering
	\includegraphics[width=\linewidth]{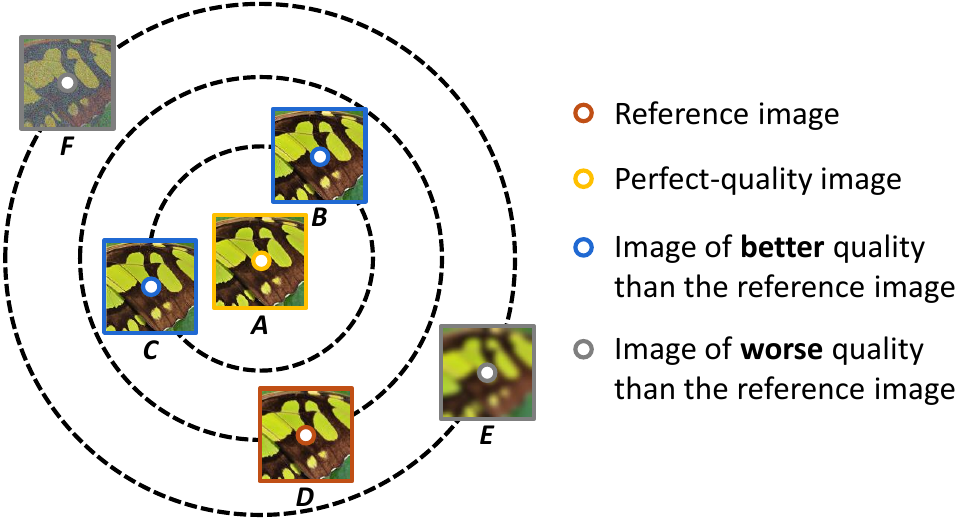}
\vspace{-2mm}
    \caption{Standard FR-IQA models tend to fail when the reference image is of non-perfect quality. In this visualization, images are embedded in a perceptually uniform space, where the perceived quality of a test image is described by its Euclidean distance to the perfect-quality image. Images located on the same dashed circles are perceived to have identical visual quality.}
	\label{fig:motivation}
 \vspace{-3mm}
\end{figure}

The violation of the perfect reference quality assumption undermines the reliability and applicability of 
standard FR-IQA models in providing useful quality estimates. \figurename~\ref{fig:motivation} shows a motivating example. We embed images in a perceptually uniform space, where the perceived quality of a test image is computed by its Euclidean distance to the perfect-quality image (\ie, \textit{Image A}). When the reference \textit{Image D} of non-perfect quality is used, it is inherently incapable of assessing \textit{Images B} and \textit{C} with higher quality. Additionally, it may also struggle to accurately evaluate \textit{Images E} and \textit{F} of the same worse quality (\ie, lying on the same level set). In this case, standard FR-IQA models may be biased toward \textit{Image E}, which is closer to the imperfect reference.

\begin{figure*}[t]
	\centering
	\includegraphics[width=\textwidth]{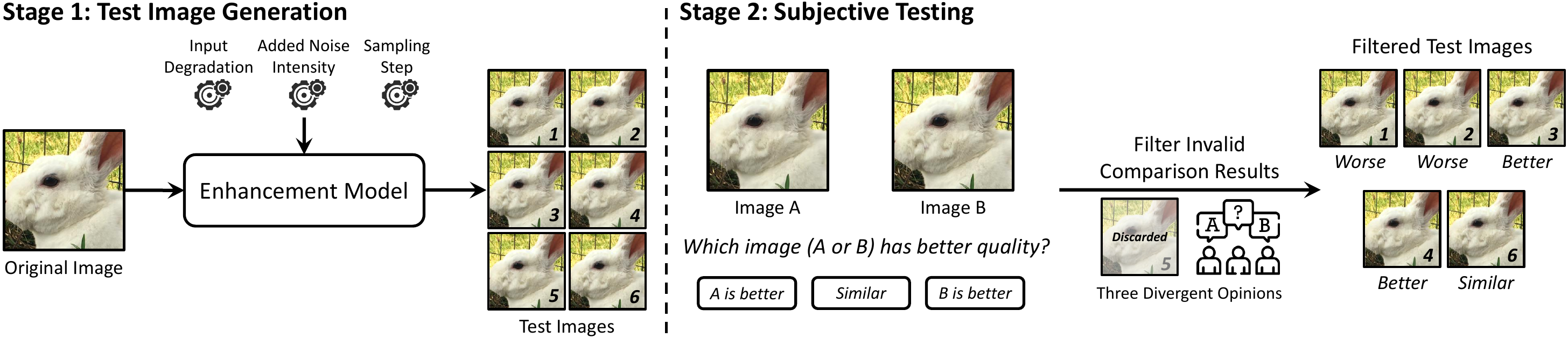}
\vspace{-2mm}
    \caption{DiffIQA is constructed in two stages. In Stage 1, we adapt PASD~\cite{yang2023pixel} to a generative image enhancer (see the {Appendix} for more details) to produce images of varying perceptual quality, some of which are perceived better than the original. In Stage 2, we conduct subjective experiments using incomplete paired comparison, followed by raw subjective data filtering.}
	\label{fig:illustration-of-DiffIQA}
    \vspace{-3mm}
\end{figure*}

Very limited subjective testing~\cite{sheikh2006image} and objective modeling~\cite{zheng2021learning} studies have been reported on the quality assessment of enhanced images under imperfect reference conditions. These studies are becoming increasingly obsolete as they primarily focused on simple and synthetic scenarios (\eg, image interpolation~\cite{yeganeh2015objective} and Gaussian image denoising~\cite{zhang2018corrupted}), and the employed enhancers frequently fail to yield images with improved perceptual quality.

To relax the perfect reference quality assumption, we first establish a large-scale IQA database, named DiffIQA, which comprises approximately $180,000$ images with \textit{worse}, \textit{similar}, and \textit{better} quality relative to their corresponding references. DiffIQA is generated by adapting the recent pixel-aware stable diffusion (PASD) method~\cite{yang2023pixel} into a powerful image enhancer, while also adjusting its hyper-parameters and perturbing the input reference images. We then invite human subjects to categorize each image as having worse, similar, or better perceptual quality compared to its reference using incomplete paired comparison. Moreover, we present a generalized FR-IQA model by adaptively weighting a DISTS-like image fidelity term and an image naturalness term, both of which share the same feature extraction backbone. The resulting \textbf{A}daptive \textbf{FI}delity-\textbf{N}aturalness \textbf{E}valuator (A-FINE) can be end-to-end optimized, and gracefully reverted to standard FR-IQA models when the reference image is much more natural than the test image. To further evaluate A-FINE, we construct an SR-based IQA benchmark, named SRIQA-Bench, comprising $1,000$ images generated by ten SR methods and annotated using complete paired comparison.

In summary, the contributions of this paper include
\begin{itemize}
	\item A large-scale IQA database, DiffIQA, breaking the perfect reference quality assumption;
	\item A generalized FR-IQA model, A-FINE, outperforming existing methods under both perfect and imperfect reference conditions;
	\item An extensive experimental demonstration on the effectiveness of A-FINE on standard IQA datasets~\cite{larson2010most, ponomarenko2015image, gu2020pipal, lin2019kadid}, and our newly created DiffIQA and SRIQA-Bench.
\end{itemize}

\section{Related Work}

\noindent\textbf{FR-IQA Datasets.} The creation of FR-IQA datasets generally starts by selecting a set of reference images of ``perfect'' quality, to which multiple synthetic distortions at various intensity levels are applied. Subjective testing is then conducted to gather mean opinion scores (MOSs) as the ground-truth quality annotations for the distorted images.

The LIVE dataset~\cite{sheikh2006statistical} is the first successful public-domain IQA dataset, containing $29$ reference images and five types of distortions, annotated using a single-stimulus continuous quality rating method. CSIQ \cite{larson2010most} maintains a similar dataset size but enhances annotation efficiency using multi-stimulus continuous quality rating. TID2013 \cite{ponomarenko2015image} extends the distortion scope to $25$ types, and utilizes an incomplete pairwise comparison method based on the
Swiss tournament system, constrained to pairs of the same underlying visual content.
The KADID-10K \cite{lin2019kadid} dataset has 81 reference images, yielding $10,125$ distorted images rated using double-stimulus absolute category rating on a crowdsourcing platform. The Waterloo Exploration Database~\cite{ma2016waterloo} expands the number of reference images to $4,744$, and introduces three computational tests for evaluating IQA models without reliance on subjective experimentation. BAPPS \cite{zhang2018unreasonable} and PIPAL \cite{gu2020pipal} broaden the scope of synthetic distortion scenarios by incorporating algorithm-dependent distortions from DNN-based image restoration and enhancement algorithms (see Table~\ref{tab:FRIQA-datasets}). 

\noindent\textbf{Standard FR-IQA for Distorted Images.} Mean squared error (MSE, along with its derivative PSNR) and mean absolute error (MAE) have been dominantly used, yet they fail to match human perception of visual quality. Within this error visibility paradigm, various remedies have been proposed,  including VDP~\cite{daly1992visible} and its HDR extensions~\cite{mantiuk2005predicting, mantiuk2011hdr}, MAD~\cite{larson2010most}, and LPIPS~\cite{zhang2018unreasonable}. A major paradigm shift occurred in 2004 with the introduction of SSIM~\cite{wang2004image}, which prioritizes structural similarity over error visibility. SSIM has been extended for multiscale processing~\cite{wang2003multiscale} and transformed/feature domain analysis~\cite{zhang2011fsim}. Leveraging pretrained DNN-based features, DISTS~\cite{ding2020image} and its locally adaptive version~\cite{ding2021locally} unify the structural and textural similarity. Standard FR-IQA models rely on comparing the test image against a perfect-quality reference image, and thus they fall short in quality assessment of enhanced images.

\noindent\textbf{Generalized FR-IQA for Enhanced Images.} Although not initially intended for this purpose, the information-theoretic VIF~\cite{sheikh2006image} is one of the first FR-IQA methods to handle cases when the test image visually outperforms the reference. PCQI \cite{wang2015patch} takes a structural similarity approach, and gives credit to image patches with improved local contrast. Yeganeh and Wang~\cite{yeganeh2015objective} leveraged the low-resolution image to evaluate interpolated image quality based on a natural scene statistical model, while Zhang \etal~\cite{zhang2018corrupted} used the noisy image to predict denoised image quality through empirical Bayes estimation. CKDN \cite{zheng2021learning} aligns the degraded image with the reference in feature space, allowing its features to act as a reference proxy for assessing restored images. These models, developed and tested under simplistic, constrained scenarios,  tend to struggle on the proposed DiffIQA and SRIQA-bench, which feature test images with improved quality compared to their corresponding references. 
The proposed A-FINE is designed as a generalized FR-IQA model, which can be end-to-end optimized to perform well under both perfect and imperfect reference conditions.

\begin{table*}[t]
\centering
\huge
\renewcommand{\arraystretch}{1.0}
\caption{Comparison of DiffIQA and SRIQA-Bench against existing representative FR-IQA datasets.}
% \scalebox{0.8}{
\resizebox{\linewidth}{!}{
\begin{tabular}{l|cccccc}
			\textbf{Dataset} & \textbf{\begin{tabular}[c]{@{}c@{}}\# of \\ Ref. Images\end{tabular}} & \textbf{\begin{tabular}[c]{@{}c@{}}\# of \\ Test Images\end{tabular}} & \textbf{\begin{tabular}[c]{@{}c@{}}Distortion /  \\ Enhancement Type\end{tabular}} & \textbf{\begin{tabular}[c]{@{}c@{}} Image \\ Resolution\end{tabular}} & \textbf{\begin{tabular}[c]{@{}c@{}}\# of Human\\ Annotations\end{tabular}} & \textbf{\begin{tabular}[c]{@{}c@{}}Perfect Reference \\ Quality Assumption\end{tabular}} \\ \Xhline{4\arrayrulewidth}
			LIVE~\cite{sheikh2006statistical} & 29  & 779  & Simulated & 480$\times$720 to 768$\times$512   & 25k & Necessary  \\
			CSIQ~\cite{larson2010most} & 30  & 866  & Simulated     & 512$\times$512 & 25k & Necessary     \\
			TID2013~\cite{ponomarenko2015image}  & 25  & 3k   & Simulated     & 512$\times$384  & 500k    & Necessary  \\
			KADID-10K~\cite{lin2019kadid}   & 81  & 10.1k  & Simulated     & 512$\times$384  & 303.8k  & Necessary \\
			PIPAL~\cite{gu2020pipal}    & 250 & 29k  & Simulated / GAN-based  & 288$\times$288   & 1.1m    & Necessary \\
			BAPPS~\cite{zhang2018unreasonable}    & 187.7k    & 375.4k & Simulated / DNN-based  & 64$\times$64    & 484.3k  & Necessary  \\ \hline
			DiffIQA (Ours) & 29.9k & 177.3k & Diffusion-based  & 512$\times$512  & 537.6k  & Not Necessary     \\
            SRIQA-Bench (Ours) & 100 & 1.1k & DNN- / GAN- / Diffusion-based & 512$\times$512  & 55k  & Not Necessary     \\
		\end{tabular}}
    \label{tab:FRIQA-datasets}
	% }
\end{table*}

\section{Proposed Dataset: DiffIQA}
\label{sec:DiffIQA}
This section describes the construction of DiffIQA, including test image generation by our diffusion-based image enhancer and subjective testing for collecting quality annotations, as illustrated in~\figurename~\ref{fig:illustration-of-DiffIQA}.

\subsection{Generative Image Enhancer}
\label{sec:Generative Image Enhancer}

To generate test images with diverse quality, we 
adapt the PASD method~\cite{yang2023pixel}, which is initially designed for realistic single-image SR and personalized stylization, into a generative image enhancer. Specifically, we feed the input image to a lightweight convolutional network to generate the control signal for the ControlNet~\cite{zhang2023adding}, and employ the pretrained Stable-Diffusion~\cite{rombach2022high} as the backbone to enhance the image. In the backward diffusion process, we incorporate pixel-aware cross-attention~\cite{yang2023pixel} to facilitate
interactions between generative features in diffusion UNet and the control features from  ControlNet.
The proposed enhancer is trained on the widely-adopted DF2K\_OST~\cite{wang2021real,wang2018recovering}, DIV8K~\cite{gu2019div8k}, FFHQ~\cite{karras2019style}, and LSDIR~\cite{li2023lsdir} datasets. To diversify output image quality, half of the input images are subject to slight blind degradations~\cite{wang2021real}, allowing the enhancer to produce outputs with \textit{worse}, \textit{similar}, and \textit{better} visual quality compared to the original. The trainable components of our enhancer---the lightweight convolutional network and the pixel-aware cross-attention modules---are optimized to predict the noise added to the input latent. More details regarding the network architecture and training procedure are presented in the {Appendix}.

\subsection{Construction of DiffIQA}
\label{sec:Construction of DiffIQA}

\noindent\textbf{Test Image Generation.} We gathered original input images from three sources: 1) $1,200$ from the DF2K dataset \cite{wang2021real}; 2) $1,000$ from the Internet under the license of Creative Commons; and 3) $640$ captured using mobile phones or digital cameras. These were cropped to $512 \times 512$ with an overlap of less than $128$ pixels, leading to a total of $29,868$ images as inputs to our trained generative enhancer.
During inference, we randomly 1) applied the same degradations as used during training, 2) augmented the initial image latent with additive Gaussian noise of varying intensities, and 3) adjusted the sampling steps within range $[20, 1000]$ to generate images with diverse quality levels.  We produced six test images for each input, totaling $179,208$ images. Additional details are provided in the {Appendix}.

\noindent\textbf{Subjective Testing.} We employed an incomplete paired comparison method, where subjects were shown a reference image alongside a test image of the same visual content in random spatial order. They were asked to infer the relative quality of the two images by choosing one from three options:
the left image is of \textit{worse}, \textit{similar}, \textit{better} perceived quality compared to the right one. A diverse group of $240$ subjects, including $132$ males and $108$ females aged between $18$ and $42$, contributed to this study. A total of $179,208$ image pairs were evaluated, with each subject assigned $2,240$ pairs, organized into multiple $30$-minute sessions to mitigate visual fatigue. All subjects completed the assigned sessions within two weeks, and the entire subjective testing spanned four months. Each image pair was rated by a minimum of three annotators, and the average time taken for each comparison was about $3.40$ seconds. More details regarding the subjective experimental setups can be found in the {Appendix}.

\begin{figure*}[t]
	\centering
	\includegraphics[width=\textwidth]{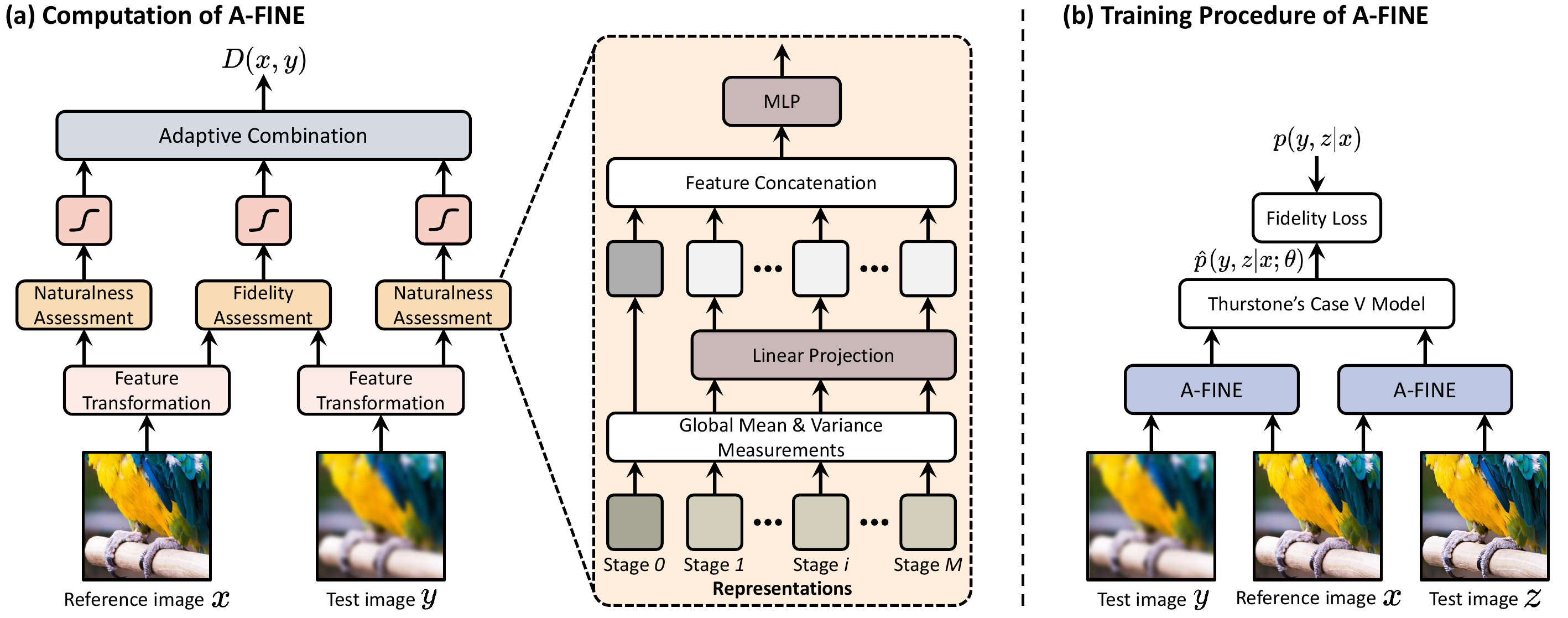}
	\caption{System diagram of the proposed A-FINE and its pairwise learning-to-rank training procedure. A-FINE leverages a shared feature transformation to make image fidelity and naturalness measurements, which are adaptively combined to produce the final quality score.}
	\label{fig:A-FINE}
\end{figure*}

\subsection{DiffIQA Statistics}
We gathered a total of $537,624$ quality annotations, with $232,285$ (43.20\%) labeled as \textit{worse}, $85,671$ (15.94\%) as \textit{similar}, and $219,668$ (40.86\%) as \textit{better} compared to the reference. This verifies the capability of our generative enhancer to improve the perceived quality of original input images. The potential labeling discrepancies among the three subjects were resolved by majority voting. When there is a tie (\ie, one \textit {worse}, one \textit{similar} and one \textit{better} vote), the image pair is marked as outlier and removed. As a result, we discard $1,889$ (1.05\%) invalid annotations, leading to  $76,515$ (42.70\%) \textit{worse}, $24,654$ (13.76\%) \textit{similar}, and $76,150$ (42.49\%) \textit{better} quality labels.

As summarized in \tablename~\ref{tab:FRIQA-datasets}, the proposed DiffIQA dataset possesses three unique features that distinguish it from existing FR-IQA datasets. First, it is large-scale in terms of the number of test images (at the standard resolution of $512\times 512$) and the number of quality annotations. Second, it features diffusion-based distortions, which exhibit visual characteristics distinct from those produced by regression-based or GAN-based image restoration and enhancement methods. Third, it is generalized to include test images with better quality than their references, relaxing the perfect reference quality assumption.

\section{Proposed FR-IQA Model: A-FINE}
In this section, we introduce the design of A-FINE as a generalized FR-IQA model, followed by a detailed description of its training procedure. The overall system diagram of A-FINE is illustrated in \figurename~\ref{fig:A-FINE}.

\subsection{Computation of A-FINE}
Given a reference image $x \in \mathbb{R}^{H\times W \times 3}$, which may be of lower quality than the test image $y\in \mathbb{R}^{H\times W\times 3}$, we aim to learn a generalized FR-IQA model, $D(x,y)$, to evaluate the perceptual quality of $y$ relative to $x$ without the perfect reference quality assumption. Drawing inspiration from  the maximum a posterior (MAP) estimation, A-FINE is designed as an adaptive linear combination of an image fidelity term, $F(x,y)$, and an image prior term, $N(y)$:
\begin{equation}
	D(x,y)=F(x,y)+\lambda(x,y)N(y),
	\label{eq:general_formulation}
\end{equation}
where $\lambda(x,y)\ge 0 $ serves as the adaptive weighting function. The value of $D(x,y)$ can either be positive or negative, with a smaller value indicating better-predicted quality of $y$. Correspondingly, smaller values of $F(x,y)$ and $N(y)$ denote better-predicted fidelity and naturalness.

Intuitively, $\lambda(x,y)$ can be designed as a function reflecting the relative naturalness of the two images. If the naturalness score of the reference image, $N(x)$, is significantly lower than that of the test image, $N(y)$, the term $F(x,y)$ should largely drive the quality prediction, thereby reducing $D(x,y)$ to be a standard FR-IQA model. Conversely, when the test image $y$ appears more natural than $x$, as in the case when $x$ is degraded, $D(x,y)$ should depend more on the naturalness assessment of the test image itself, \ie, $N(y)$. 
Thus, we define $\lambda(x,y)$ as
\begin{equation}
	\lambda(x,y)=\exp\left(k(N(x)-N(y)\right),
	\label{eq:lambda}
\end{equation}
where $k\ge 0$ is a learnable scale parameter.

To instantiate the image fidelity term $F(x,y)$, we employ a DISTS-like \cite{ding2020image, ding2021locally} approach:
\begin{equation}
	F(x,y)=1-\sum_{i=0}^{M} \sum_{j=1}^{N_{i}}F\left(x_{j}^{(i)}, y_{j}^{(i)}\right),
	\label{eq:D(x,y)}
\end{equation}
where
\begin{equation}
	F\left(x_{j}^{(i)}, y_{j}^{(i)}\right)=\alpha_{i j} L\left(x_{j}^{(i)}, y_{j}^{(i)}\right)+\beta_{i j} S\left(x_{j}^{(i)}, y_{j}^{(i)}\right).
	\label{eq:Fij}
\end{equation}
Here, $x_{j}^{(i)}$ and $y_{j}^{(i)}$ represent the feature maps extracted from the $j$-th channel of the $i$-th stage of the backbone network, corresponding to $x$ and $y$, respectively. $M$ and $N_i$ denote the total number of stages and the number of channels in the $i$-th stage, respectively. $L\left(x_{j}^{(i)}, y_{j}^{(i)}\right)$ and $S\left(x_{j}^{(i)}, y_{j}^{(i)}\right)$ measure the global texture and structure similarity~\cite{wang2004image}, respectively:
\begin{align}
	L({x}^{(i)}_j,{y}^{(i)}_j)=\frac{2 \mu_{{x}_j}^{(i)}\mu_{{y}_j}^{(i)}+c_{1}}{\left(\mu_{{x}_j}^{(i)}\right)^{2}+\left(\mu_{{y}_j}^{(i)}\right)^{2}+c_{1}},
	\label{eq:s1}
\end{align}
\begin{align}
	S({x}^{(i)}_j,{y}^{(i)}_j)=\frac{2 \sigma_{{x}_j{y}_j}^{(i)}+c_{2}}{\left(\sigma_{{x}_j}^{(i)}\right)^{2}+\left(\sigma_{{y}_j}^{(i)}\right)^{2}+c_{2}},
	\label{eq:s2}
\end{align}
where $\mu_{{x}_j}^{(i)}$, $\mu_{{y}_j}^{(i)}$, $(\sigma_{{x}_j}^{(i)})^{2}$, $(\sigma_{{y}_j}^{(i)})^{2}$, and $\sigma_{{x}_j{y}_j}^{(i)}$ denote the global means and variances of ${x}^{(i)}_j$ and ${y}^{(i)}_j$, as well as the global covariance between ${x}^{(i)}_j$ and ${y}^{(i)}_j$, respectively. $c_1$ and $c_2$ are two small positive constants to prevent numerical instability when the denominators approach zero. The weights $\{\alpha_{i j}, \beta_{i j}\}$ are positive and learnable, satisfying the constraint $\sum_{i=0}^{M} \sum_{j=1}^{N_{i}} \left(\alpha_{i j} + \beta_{i j}\right) = 1$. 

In contrast to DISTS~\cite{ding2020image}, which uses a fixed, pretrained VGG network~\cite{simonyan2014very} for feature representation, A-FINE adopts a more advanced Vision Transformer (ViT) as the backbone, specifically the CLIP ViT-B/$32$@$224$~\cite{radford2021learning}. We interpolate position embeddings to accommodate images of arbitrary resolutions. Additionally, we fine-tune all backbone parameters, denoted as $\phi$, along with the linear weights in Eq.~\eqref{eq:Fij}, exploiting the transferability of ViT features to improve quality prediction.

To instantiate the image naturalness term $N(\cdot)$, parameterized by $\varphi$, we reuse the CLIP ViT backbone for computing $F(x,y)$. Global average and variance pooling are applied to the feature maps at each stage, resulting in a stage-wise feature vector of size $768 \times 2$. This vector is then linearly projected into a $128$-dimensional space using a shared projection matrix across all stages. The projected multi-stage features are 
concatenated and fed to a multilayer perceptron (MLP), composed of two fully connected layers and a Gaussian error linear unit (GELU) activation in between, with dimensions $3 \times 2 + 128 \times 12 \rightarrow 128\times 6 \rightarrow 1 $, to compute the naturalness score\footnote{The term ``$3\times 2$'' corresponds to the global means and variances computed from the three color channels of the input test image.}. 

To stabilize training, we follow~\cite{sheikh2006statistical} and incorporate a separate four-parameter monotonic logistic function for both $F(x,y)$ and $N(y)$ as part of our model computation:

\begin{equation}
    F_\eta(x,y)= \frac{\eta_1 - \eta_2}{1 + \exp\left(-\frac{F(x,y)-\eta_3}{\vert\eta_4\vert}\right)} + \eta_2
    \label{eq:non-linear F}
\end{equation}
and
\begin{equation}
    N_\gamma(y)= \frac{\gamma_1 - \gamma_2}{1 + \exp\left(-\frac{N(y)-\gamma_3}{\vert\gamma_4\vert}\right)} + \gamma_2,
    \label{eq:non-linear N}
\end{equation}
where $\eta_1$ and $\gamma_1$ are set to $2$, and $\eta_2$ and $\gamma_2$ are set to $-2$, respectively, defining the upper and lower bounds of the non-linear mappings. The learnable parameters constitute $\{\eta_3, \eta_4, \gamma_3, \gamma_4\}$.

\subsection{Training Procedure of A-FINE}
\label{Training Procedure of A-FINE}
Inspired by UNIQUE~\cite{zhang2021uncertainty}, we adopt a similar pairwise learning-to-rank approach to optimize the parameters in A-FINE, collectively denoted as $\theta = \{\phi, \varphi, \alpha, \beta, k, \eta, \gamma\}$. In particular, given a triplet $(x, y, z)$, where $x$ is the reference image, and $y$ and $z$ are two test images with the same underlying visual content as $x$, we derive the ground-truth ranking label based on the relative quality of $y$ and $z$:
\begin{equation}
	{p}(y, z|x) =
	\begin{cases}
		1 & \text{if } Q (y|x) > Q (z|x) \\
		0.5 & \text{if } Q(y|x) = Q(z|x) \\
		0 & \text{otherwise},
	\end{cases}
	\label{eq:label}
\end{equation}
where ${Q}(y|x)$ and $Q(z|x)$ represent the MOS of $y$ and $z$ relative to $x$, respectively. Under Thurstone's Case V model~\cite{thurstone1927law}, we assume that the perceptual quality of a test image follows a Gaussian distribution, where the mean is estimated by the proposed A-FINE, and the variance is fixed to one. This enables us to compute the probability that $y$ is perceived better than $z$ given $x$ by:
\begin{equation}
	\hat{p}(y, z|x;\theta)=\Phi\left(\frac{D(x,y;\theta)-D(x, z;\theta)}{\sqrt{2}}\right), \label{eq}
\end{equation}
where $\Phi(\cdot)$ represents the standard Gaussian cumulative distribution function. Following~\cite{zhang2021uncertainty}, we adopt the fidelity loss~\cite{tsai2007frank} for end-to-end optimization:
\begin{equation}
	\begin{aligned}
		\ell(y, z|x;\theta) & =1-\sqrt{p(y, z|x) \hat{p}(y, z|x;\theta)} \\
		& -\sqrt{(1-p(y, z|x))(1-\hat{p}(y, z|x;\theta))}.
	\end{aligned}
	\label{eq:fidelity_loss}
\end{equation}

\begin{table*}[t]
\centering
\large
\renewcommand{\arraystretch}{1.0}
\caption{Accuracy (\%) results of FR-IQA models on the test sets of TID2013~\cite{ponomarenko2015image}, KADID-10K~\cite{lin2019kadid}, PIPAL~\cite{gu2020pipal}, and the proposed DiffIQA. The term ``Combined'' indicates the combination of TID2013, KADID, PIPAL, and DiffIQA. The suffix ``-FT'' means that the model is fine-tuned on this combined dataset. The top two results are highlighted in \textbf{bold} and \underline{underlined}, respectively.}
\resizebox{\linewidth}{!}{
\begin{tabular}{cll|cccc|ccc|c}
\multirow{2}{*}{\textbf{Scenario}}    & \multirow{2}{*}{\textbf{Method}} & \multirow{2}{*}{\textbf{\begin{tabular}[c]{@{}l@{}}Training\\ Dataset\end{tabular}}} & \multirow{2}{*}{\textbf{TID2013}} & \multirow{2}{*}{\textbf{KADID}} & \multirow{2}{*}{\textbf{PIPAL}} & \multirow{2}{*}{\textbf{Average}} & \multicolumn{3}{c|}{\textbf{DiffIQA}} & \multirow{2}{*}{\textbf{\begin{tabular}[c]{@{}c@{}}All\\ Average\end{tabular}}} \\
 & & &  &   &   & & \textbf{Ref} $\mathbf{<}$ \textbf{Test} & \textbf{Ref} $\mathbf{>}$ \textbf{Test} & \textbf{Average}  &    \\ \Xhline{3\arrayrulewidth}
\multirow{13}{*}{Standard}   & PSNR    & N.A.    & 75.8     & 74.8   & 70.7   & 72.2 & 18.2   & 92.1  & 45.6 & 58.9   \\
 & SSIM~\cite{wang2004image}    & N.A.    & 68.9     & 74.0   & 72.1   & 72.4 & 20.1   & 93.0  & 47.1 & 60.0   \\
 & MS-SSIM~\cite{wang2003multiscale} & N.A.    & 83.4     & 81.8   & 72.5   & 75.9 & 20.1   & 93.0  & 47.1 & 61.5   \\
 & FSIM~\cite{zhang2011fsim}    & N.A.    & 86.0     & 83.4   & 76.2   & 79.0 & 20.2   & 93.1  & 47.2 & 63.1   \\
 & VSI~\cite{zhang2014vsi}     & N.A.    & 87.3     & 84.8   & 76.2   & 79.5 & 19.7   & 93.1  & 46.9 & 63.2   \\
 & LPIPS~\cite{zhang2018unreasonable}   & BAPPS   & 78.7     & 77.0   & 74.3   & 75.4 & 23.7   & 94.7  & 50.0 & 62.7   \\
 & LPIPS-FT    & Combined    & 72.5     & 78.2   & 71.7   & 73.6 & 35.4   & 91.6  & 55.6 & 64.6   \\
 & DISTS~\cite{ding2020image}   & KADID   & 78.4     & 81.4   & 75.3   & 77.2 & 21.4   & \underline{94.8}  & 48.6 & 62.9   \\
 & DISTS-FT    & Combined    & 78.4     & 81.9   & 72.1   & 75.3 & 38.2   & 89.5  & 56.7 & 66.0   \\
 & AHIQ~\cite{lao2022attentions}    & PIPAL   & 74.6     & 76.4   & 79.3   & 78.1 & 34.1   & 88.1  & 54.1 & 66.1   \\
 & AHIQ-FT & Combined    & 81.0     & 79.7   & 74.9   & 76.7 & 78.4   & 73.8  & 76.7 & 76.7   \\
 & TOPIQ~\cite{chen2024topiq}   & KADID   & \textbf{90.4}     & \textbf{94.3}   & \underline{80.5}   & \textbf{85.1} & 22.1   & \textbf{95.1}  & 49.1 & 67.1   \\
 & TOPIQ-FT    & Combined    & 78.9     & 85.0   & 79.0   & 80.6 &\underline{78.6}   & 74.2  & \underline{77.0}   & \underline{78.8}   \\ \hline
\multirow{6}{*}{Generalized} & VIF~\cite{sheikh2006image}     & N.A.    & 78.5     & 75.2   & 72.4   & 73.7 & 20.0   & 92.8  & 46.9 & 60.3   \\
 & PCQI~\cite{wang2015patch}    & N.A.    & 66.6     & 65.4   & 56.7   & 59.9 & 17.3   & 90.3  & 44.3   & 52.1   \\
 & SFSN~\cite{zhou2022quality}    & N.A.    & 75.6     & 70.5   & 69.8   & 70.5 & 19.5   & 89.6  & 45.4 & 58.0   \\
 & CKDN~\cite{zheng2021learning}    & PIPAL   & 76.9     & 70.9   & 79.8   & 77.1 & 33.3   & 82.4  & 51.4 & 64.3   \\
 & CKDN-FT & Combined    & 75.0     & 80.1   & 68.1   & 72.0 & \textbf{79.4}   & 71.0  & 76.4 & 74.2   \\ \cline{2-11}
 & A-FINE (Ours)  & Combined    & \underline{88.1}     & \underline{88.3}   & \textbf{81.0}   & \underline{83.6} & 78.5  & 82.3  & \textbf{79.9} & \textbf{81.8}  
\end{tabular}
		}
	\label{tab:in-domain testing results}
\end{table*}

\section{Experiments}
In this section, we first describe the experimental setups, followed by the construction of SRIQA-Bench. Then, we compare A-FINE against several existing FR-IQA models across various IQA datasets, including the proposed DiffIQA and SRIQA-Bench.

\subsection{Experimental Setups}
\noindent\textbf{Training Details of A-FINE.}
Since our DiffIQA dataset is designed for scenarios where test images can have equal or higher quality than the reference, we combine it with  TID2013~\cite{ponomarenko2015image}, KADID-10K~\cite{lin2019kadid}, and PIPAL~\cite{gu2020pipal} to enhance the training of A-FINE. Following standard practice, we partition each dataset into training, validation, and test sets in a 7:1:2 ratio, ensuring content independence. 
The training of A-FINE proceeds in three phases. In Phase 1, we perform a warm-up training for the image naturalness term $N(\cdot)$, in which we fine-tune the ViT backbone parameters $\phi$, and train the linear projection matrix and the MLP prediction head, with parameters $\varphi$. In Phase 2, the fine-tuned ViT backbone is frozen, and the training focuses on optimizing the linear weights $\{\alpha_{ij}, \beta_{ij}\}$ associated with the fidelity term $F(\cdot, \cdot)$. In Phase 3, the complete A-FINE model is refined by optimizing the scale parameter $k$ in the adaptive weighting function $ \lambda(\cdot, \cdot)$ and the parameters $ \{\eta_i, \gamma_i\}$ in the two nonlinear mappings, while keeping all other fixed.

Training is carried out by employing AdamW~\cite{loshchilov2017decoupled} as the optimizer, with a weight decay factor of $10^{-3}$ and initial learning rates of $5\times 10^{-6}$ for Phase $1$, $5\times 10^{-4}$ for Phase $2$ and $1\times 10^{-3}$ for Phase $3$, subject to cosine annealing scheduling with a period of $10,000$ iterations. The training minibatch size on a single GPU is set to $128$, and we train A-FINE on four NVIDIA V100 GPUs. The maximum number of training iterations are set to $40,000$, $40,000$, and $10,000$ for Phases $1$, $2$, and $3$, respectively.

\noindent\textbf{Competing FR-IQA Models.} We compare A-FINE against nine standard FR-IQA models: 1) PSNR, 2) SSIM \cite{wang2004image}, 3) MS-SSIM \cite{wang2003multiscale}, 4) FSIM \cite{zhang2011fsim}, 5) VSI \cite{zhang2014vsi}, 6) LPIPS \cite{zhang2018unreasonable}, 7) DISTS \cite{ding2020image}, 8) AHIQ \cite{lao2022attentions} and 9) TOPIQ \cite{chen2024topiq}, and four generalized FR-IQA models for enhanced images: 10) VIF~\cite{sheikh2006image}, 11) PCQI \cite{wang2015patch}, 12) SFSN \cite{zhou2022quality} and 13) CKDN \cite{zheng2021learning}. For a more fair comparison, we present the performance of the original and, when applicable, the fine-tuned versions (indicated by the suffix ``-FT'') of these models on the same combined dataset used to train A-FINE.

\subsection{SRIQA-Bench}
\label{sec:SRIQA-Bench}
To further verify the effectiveness of A-FINE, we constructed an SR-based IQA benchmark, named \textbf{SRIQA-Bench}. 
We first compiled $100$ original images covering a wide range of natural scenes and subjected them to common degradations~\cite{wang2021real, zhang2021designing} to generate input low-resolution images. We then adopted two regression-based SR methods: 1) SwinIR~\cite{liang2021swinir} and 2) RRDB~\cite{wang2018esrgan}, and eight generation-based SR methods: 3) Real-ESRGAN~\cite{wang2021real}, 4) BSRGAN~\cite{zhang2021designing}, 5) HGGT~\cite{chen2023human}, 6) SUPIR~\cite{yu2024scaling}, 7) SeeSR~\cite{wu2024seesr}, 8) StableSR~\cite{wang2024exploiting}, 9) SinSR~\cite{wang2024sinsr} and 10) OSEDiff~\cite{wu2024one} to produce ten SR images for each input. Generally speaking, diffusion-based SR methods outperform GAN-based methods with more plausible textures, which in turn are more effective than regression-based SR methods with more realistic and sharper structures.

The subjective testing protocol is identical to the one described in Sec.~\ref{sec:DiffIQA}, except that we performed a complete paired comparison experiment involving $\binom{11}{2}=55$ pairs per input. To ensure rating reliability, each pair was assessed by at least ten subjects. A total of $40$ subjects---comprising $25$ males and $15$ females aged between $21$ and $39$---took part in this study. More details about dataset construction can be found in the {Appendix}.

\subsection{Main Results}

\noindent\textbf{Within-Dataset Results.} Table~\ref{tab:in-domain testing results} shows the accuracy results on the test sets of TID2013~\cite{ponomarenko2015image}, KADID-10K~\cite{lin2019kadid}, PIPAL~\cite{gu2020pipal}, and the proposed DiffIQA, from which we have several key observations. First, existing FR-IQA models, standard or generalized, exhibit noticeable performance declines on DiffIQA when the assumption of perfect reference quality is not met. Second,  models based on \textit{surjective} feature transformations (\eg, AHIQ~\cite{lao2022attentions}, TOPIQ~\cite{chen2024topiq}, and CKDN~\cite{zheng2021learning}) demonstrate more pronounced improvements after fine-tuning on the combined dataset compared to models based on \textit{injective} transformations (\eg, LPIPS~\cite{zhang2018unreasonable} and DISTS~\cite{ding2020image}). This suggests that DiffIQA is beneficial for enhancing generalized FR-IQA. Nonetheless, fine-tuned models typically show a performance drop on standard IQA datasets. Last, A-FINE achieves the highest average results, attributed to its adaptive weighting of image fidelity and naturalness terms.

\begin{table}[t]
\centering
\renewcommand{\arraystretch}{1.0}
\caption{Accuracy (\%) results on SRIQA-Bench. Pairs are formed within each subcategory and across the entire dataset.}
\vspace{-2mm}
\resizebox{1.0\linewidth}{!}{
\begin{tabular}{lccc}
\textbf{Method}   & \textbf{Regression-based}    & \textbf{Generation-based}   & \textbf{All}   \\ \Xhline{2\arrayrulewidth}
PSNR     & 80.7  & 41.7  & 34.7  \\
SSIM~\cite{wang2004image}    & 83.0  & 45.3  & 37.4  \\
MS-SSIM~\cite{wang2003multiscale}  & 83.0  & 45.6  & 37.6  \\
FSIM~\cite{zhang2011fsim}    & \underline{85.3} & 49.5  & 41.0  \\
VSI~\cite{zhang2014vsi}    & 81.3  & 50.1  & 41.2  \\
LPIPS~\cite{zhang2018unreasonable}   & 82.0  & 63.9  & 65.8  \\
LPIPS-FT & 84.7  & 63.8  & 72.2  \\
DISTS~\cite{ding2020image}   & 83.3  & 66.6  & 72.4  \\
DISTS-FT & \textbf{86.0} & 63.9  & 71.4  \\
AHIQ~\cite{lao2022attentions}   & 83.7  & 70.0  & 68.4  \\
AHIQ-FT  & 71.0  & 71.5  & 69.6  \\
TOPIQ~\cite{chen2024topiq}    & 83.7  & 63.9  & 67.0  \\
TOPIQ-FT & 78.3  & \underline{73.0} & \underline{77.7} \\ \hline
VIF~\cite{sheikh2006image}     & \underline{85.3} & 47.1  & 39.0  \\
PCQI~\cite{wang2015patch}   & 79.0  & 39.8  & 32.2  \\
SFSN~\cite{zhou2022quality}     & 80.3  & 48.4  & 39.9  \\
CKDN~\cite{zheng2021learning}     & 45.0  & 60.1  & 47.4  \\
CKDN-FT  & 76.7  & 64.3  & 59.1  \\ \hline
A-FINE (Ours)   & 83.3  & \textbf{78.9} & \textbf{82.4}
\end{tabular}
}
\label{tab:cross-datasets}
\vspace{-2mm}
\end{table}

\noindent\textbf{Results on SRIQA-Bench.} Table~\ref{tab:cross-datasets} presents the accuracy results on SRIQA-Bench. Most models perform adequately for regression-based SR methods. This is mainly because they have limited capabilities in creating plausible structures and textures. Consequently, the resulting SR images exhibit clearly inferior quality compared to their references. In contrast, generation-based SR methods can produce output images of much higher quality, challenging the perfect reference quality assumption. Fine-tuned models on DiffIQA show clearly improved performance, and the proposed A-FINE demonstrates the strongest generalization to SRIQA-Bench, confirming its effectiveness in evaluating enhanced image quality.

\subsection{Ablation Studies}

\noindent\textbf{Backbone.} We try different backbone networks to implement A-FINE, including VGG16~\cite{simonyan2014very}, ImageNet-trained ResNet50~\cite{he2016deep}, CLIP-trained ResNet50~\cite{radford2021learning}, ImageNet-trained ViT-B/32~\cite{dosovitskiy2020image}, and CLIP-trained ViT-B/32~\cite{radford2021learning}. From the results in Table~\ref{tab:backbone}, it is evident that A-FINE benefits from more sophisticated computation (global attention over local convolution) and stronger backbone pretrained on more data (0.4B image-text pairs over 1M images).

\noindent\textbf{Training Dataset.} We train A-FINE only on DiffIQA or only on the combined standard IQA datasets TID2013~\cite{ponomarenko2015image}, PIPAL~\cite{gu2020pipal} and KADID-10k~\cite{lin2019kadid}.  Table~\ref{tab:cross datasets training} lists the results, from which we find that training on the combined DiffIQA and standard IQA datasets using the pairwise learning-to-rank approach yields the best average performance with the strongest cross-dataset generalization.

\noindent\textbf{Training Strategy.} We last compare our three-phase training strategy with a baseline that trains all parameters of A-FINE simultaneously in a single phase. As shown in Table~\ref{tab:training strategy}, single-phase training proves less effective in optimizing individual image fidelity and naturalness terms and adaptively balancing these two. In contrast, our proposed three-phase training strategy stabilizes the training dynamics, resulting in improved generalization.

\begin{table}[t]
	\centering
	\large
	\caption{Ablation study on backbone networks. The accuracy values in the ``Standard'' column are  averaged across the test sets of TID2013~\cite{ponomarenko2015image}, KADID-10K~\cite{lin2019kadid}, and PIPAL~\cite{gu2020pipal}. The results of TOPIQ-FT are included for reference. }
    \vspace{-3mm}
	\resizebox{1.0\linewidth}{!}{
\begin{tabular}{lccccc}
\multirow{2}{*}{\textbf{Backbone}} & \multirow{2}{*}{\textbf{\begin{tabular}[c]{@{}c@{}}Standard\end{tabular}}} & \multirow{2}{*}{\textbf{\begin{tabular}[c]{@{}c@{}}DiffIQA\end{tabular}}} & \multicolumn{3}{c}{\textbf{SRIQA-Bench}}  \\
   &   &     & \textbf{Reg.} & \textbf{Gen.} & \textbf{All}  \\ \Xhline{3\arrayrulewidth}
TOPIQ-FT     & 80.6   & 77.0     & 78.3     & 73.0     & 77.7     \\ \hline
VGG16     & 77.6   & 77.0     & 79.0     & 75.0     & 79.8     \\
ResNet50 (ImageNet) & 74.8   & 69.6     & \underline{84.7}  & 70.7     & 77.2     \\
ResNet50 (CLIP)     & 76.1   & 71.1     & \textbf{85.2}  & 70.3     & 75.6     \\
ViT-B/32 (ImageNet)   & \underline{81.0}     & \underline{77.7}  & 81.3     & \underline{75.5}  & \underline{80.4} \\
ViT-B/32 (CLIP) & \textbf{83.6}     & \textbf{79.9}  & 83.3     & \textbf{78.9}  & \textbf{82.4}
\end{tabular}
	}
	\label{tab:backbone}
    \vspace{-2mm}
\end{table}

\begin{table}[t]
	\centering
	\scriptsize
	\caption{Ablation study on training datasets.}
    \vspace{-3mm}
	\resizebox{1.0\linewidth}{!}{
		\begin{tabular}{lccccc}
\multirow{2}{*}{\textbf{\begin{tabular}[c]{@{}l@{}}Training\\ Dataset\end{tabular}}} & \multirow{2}{*}{\textbf{\begin{tabular}[c]{@{}c@{}}Standard\end{tabular}}} & \multirow{2}{*}{\textbf{\begin{tabular}[c]{@{}c@{}}DiffIQA\end{tabular}}} & \multicolumn{3}{c}{\textbf{SRIQA-Bench}}     \\
  &     &  & \textbf{Reg.} & \textbf{Gen.} & \textbf{All}  \\ \Xhline{2\arrayrulewidth}
Standard   & \textbf{84.1}   & 65.6  &\textbf{86.7}   & 71.8   & \underline{78.7}     \\
DiffIQA   &70.6     &\underline{79.6}  &78.3   &\underline{72.9}   &76.0  \\ 
Combined  & \underline{83.6}   & \textbf{79.9}  & \underline{83.3}  & \textbf{78.9}  & \textbf{82.4}
\end{tabular}
	}
	\label{tab:cross datasets training}
    \vspace{-2mm}
\end{table}

\begin{table}[t]
	\centering
	\scriptsize
	\caption{Ablation study on training strategies.}
    \vspace{-3mm}
	\resizebox{1.0\linewidth}{!}{
		\begin{tabular}{lccccc}
\multirow{2}{*}{\textbf{\begin{tabular}[c]{@{}l@{}}Training\\ Strategy\end{tabular}}} & \multirow{2}{*}{\textbf{\begin{tabular}[c]{@{}c@{}}Standard\end{tabular}}} & \multirow{2}{*}{\textbf{\begin{tabular}[c]{@{}c@{}}DiffIQA\end{tabular}}} & \multicolumn{3}{c}{\textbf{SRIQA-Bench}}      \\
   & &   & \textbf{Reg.} & \textbf{Gen.} & \textbf{All}  \\ \Xhline{2\arrayrulewidth}
Single-Phase   & 79.7     & 79.6    & 79.3          & 75.1          & 77.9          \\
Three-Phase & \textbf{83.6}  & \textbf{79.9}       & \textbf{83.3} & \textbf{78.9} & \textbf{82.4}
\end{tabular}
	}
	\label{tab:training strategy}
    \vspace{-2mm}
\end{table}

\section{Conclusion and Limitations}
We explored the problem of generalized FR-IQA, which relaxes the assumption of perfect reference image quality. From a data perspective, we introduced DiffIQA and SRIQA-bench to train and test generalized FR-IQA models, respectively. From a model perspective, we developed A-FINE, which adaptively combines an image fidelity term and an image naturalness term. We hope our data and model will inspire researchers in related fields to engage with the important research topic of generalized FR-IQA in the era of deep generative models.

\noindent\textbf{Limitations.} A-FINE represents one of the early efforts in generalized FR-IQA, paving the way for several promising research avenues. First, it can perform effectively when the perceptual quality of the reference image is reasonably high, even if not perfect. However, if the reference image quality is poor, the image fidelity term in A-FINE could introduce significant bias in quality prediction, which is worth deep investigation. Second, A-FINE can be viewed as a specific instance within the broader family of \textit{asymmetric} distance measures. Exploring the optimal functional form for such measures in the context of generalized FR-IQA remains an open question. Third, we empirically observed a performance trade-off when evaluating the quality of distorted versus enhanced images. Identifying the best trade-off from both data and model perspectives presents an intriguing direction for future exploration.

\section*{Acknowledgments}
This work was partly supported by the Hong Kong ITC Innovation and Technology Fund (9440379 and 9440390), and fully supported by the PolyU-OPPO Joint Innovative Research Center. We sincerely thank the volunteers who participated in our subjective study. A Human Subjects Ethics Committee approved the study and all participants signed consent forms beforehand.

{
	\small
	\bibliographystyle{ieeenat_fullname}
	\bibliography{main}
}

\clearpage

\onecolumn
\appendix

\section*{Appendix}

\noindent In this appendix, we provide the following material:
\begin{itemize} 
    \item Training details of our generative image enhancer (please refer to Sec.~\ref{sec:Generative Image Enhancer} of the main paper);
    \item Inference details of our generative image enhancer (please refer to Sec.~\ref{sec:Construction of DiffIQA} of the main paper);
    \item Details of subjective experimental setups for constructing DiffIQA (please refer to Sec.~\ref{sec:Construction of DiffIQA} of the main paper);
    \item Details of subjective experimental setups for constructing SRIQA-Bench (please refer to Sec.~\ref{sec:SRIQA-Bench} of the main paper);
    \item Discussions on the generated ``fake'' details.
\end{itemize}

\section{Training Details of the Generative Image Enhancer}

\subsection{Architecture}

\begin{figure}[h]
	\centering
	\includegraphics[width=0.7\linewidth]{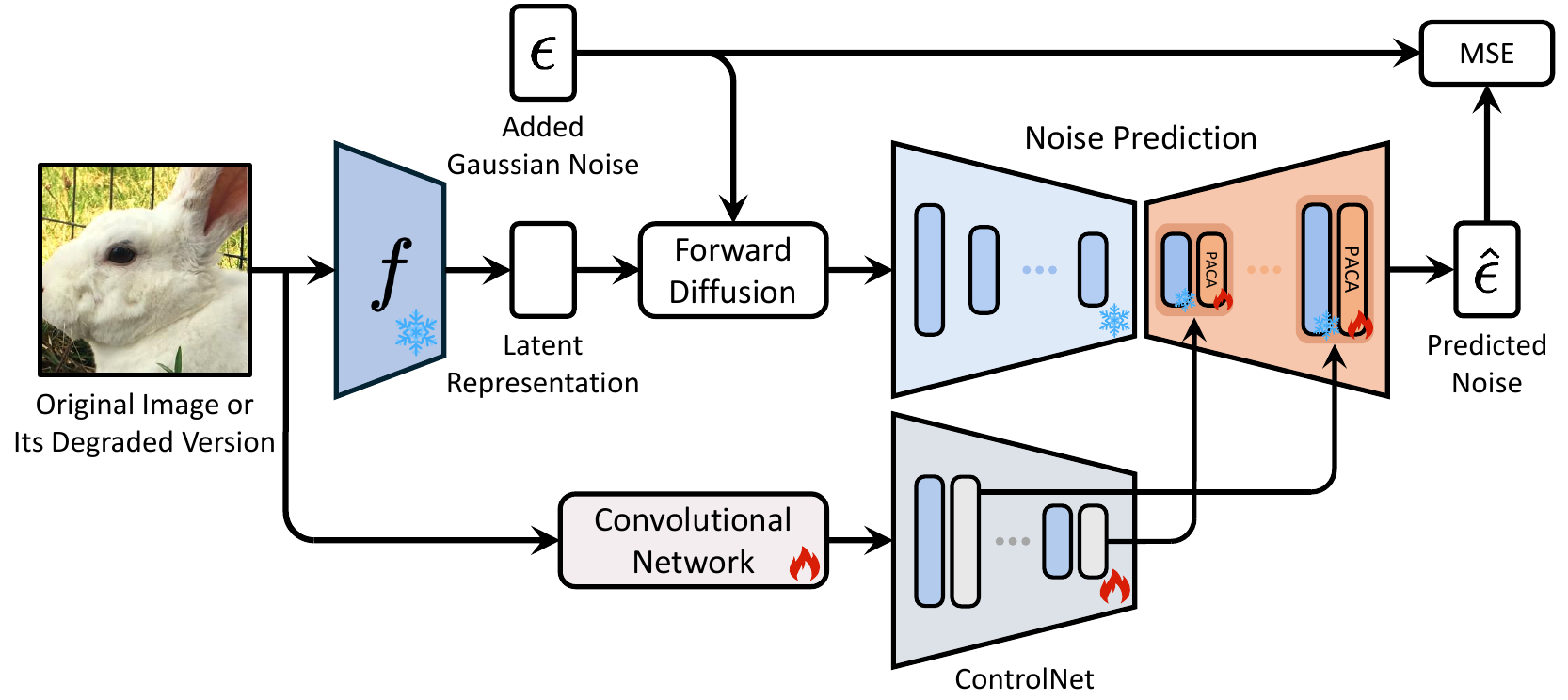}
    \caption{Training of our generative image enhancer.}
	\label{fig:training_arch}
\end{figure}

The overall architecture of the enhancer during the training phase is illustrated in Fig.~\ref{fig:training_arch}. The original image is passed through a lightweight convolutional network, with features fed into a ControlNet~\cite{zhang2023adding} to provide content-aware conditional signal to the diffusion UNet~\cite{rombach2022high}. Meanwhile, the original image or its degraded version is passed through the image encoder~\cite{esser2021taming} to produce a latent representation. In forward diffusion, Gaussian noise is added to the latent image, which serves as the input to the diffusion UNet. The conditional signal from ControlNet interacts with the diffusion UNet via  pixel-aware cross-attention (PACA)~\cite{yang2023pixel}. Finally, we compute the MSE between the predicted noise by the diffusion UNet and the added Gaussian noise, which is treated as the ground-truth. During training, only the convolutional network, ControlNet, and PACA modules are trainable.

\subsection{Training Specifications}

To diversify output image quality, $50\%$ of the original inputs are directly fed into the enhancer, while the other $50\%$ undergo slight blind degradations~\cite{wang2021real, zhang2021designing}:
\begin{equation}
\label{eq:blind degradation model}
    x_d = \mathtt{compression}\left(\mathtt{resizing}(x * h) + \epsilon\right),
\end{equation}
where $x$ represents the original high-quality image, and $h$ is an (an)isotropic blur kernel. $\mathtt{resizing}(\cdot)$ indicates the resizing operation, $\epsilon$ denotes the additive Gaussian or Poisson noise, and $\mathtt{compression}(\cdot)$ stands for JPEG compression. The degraded image $x_d$ is resized to the original resolution using bicubic interpolation before feeding into the enhancer. Detailed degradation configurations are provided in Table~\ref{tab:blind degradation settings in training enhancer}.

We trained our enhancer on eight NVIDIA V100 GPUs using Adam with a fixed learning rate of $5\times 10^{-5}$ for $100,000$ iterations, each GPU handling a minibatch size of $32$.
The training image size was configured at $512\times 512$ pixels. 

\begin{table}[h]
    \centering
    \scriptsize
    \caption{Blind degradation settings of our enhancer.  ``iso'' and ``an-iso'' denote ``isotropic'' and  ``an-isotropic,'' respectively.}
    \resizebox{\linewidth}{!}{
    \begin{tabular}{c|l|c} 
        \textbf{Operation}               & \textbf{Parameter}                   & \textbf{Setting}                     \\ 
        \Xhline{2\arrayrulewidth}
        \multirow{5}{*}{Blurring}   & Kernel size $[2m+1]$          & $m \in [1, 4]$              \\
        & Kernel list                 & iso, an-iso, generalized iso, generalized an-iso, plateau iso, plateau an-iso     \\
        & Kernel list probability     & $0.45$, $0.25$, $0.12$, $0.03$, $0.12$, $0.03$                    \\
        & Sinc kernel~\cite{wang2021real} probability     & $0.1$                        \\
        & Standard deviation &  $[0.0, 1.2]$   \\ 
        \hline
        \multirow{4}{*}{Resizing} & Resizing list                 & down-sampling, up-sampling    \\
        & Resizing list probability     & $0.85$, $0.05$, $0.1$             \\
        & Resizing range         & $[0.8, 1.1]$     \\
        & Resizing mode                 & area, bilinear, bicubic     \\ 
        \hline
        \multirow{5}{*}{Noise Contamination}  & Noise list                  & Gaussian, Poisson           \\
        & Noise list probability      & $0.5$, $0.5$                    \\
        & Sigma of Gaussian  &  $[0.0, 13.0]$      \\
        & Scale of Poisson  & $[0.0, 0.9]$  \\
        & Gray noise~\cite{wang2021real} probability      & $0.1$                         \\ 
        \hline
        JPEG  Compression                  & Quality factor              & $[75, 95]$                  
    \end{tabular}
    }
    \label{tab:blind degradation settings in training enhancer}
\end{table}

\section{Inference Details of the Generative Image Enhancer}

\begin{figure}[t]
	\centering
	\includegraphics[width=0.8\linewidth]{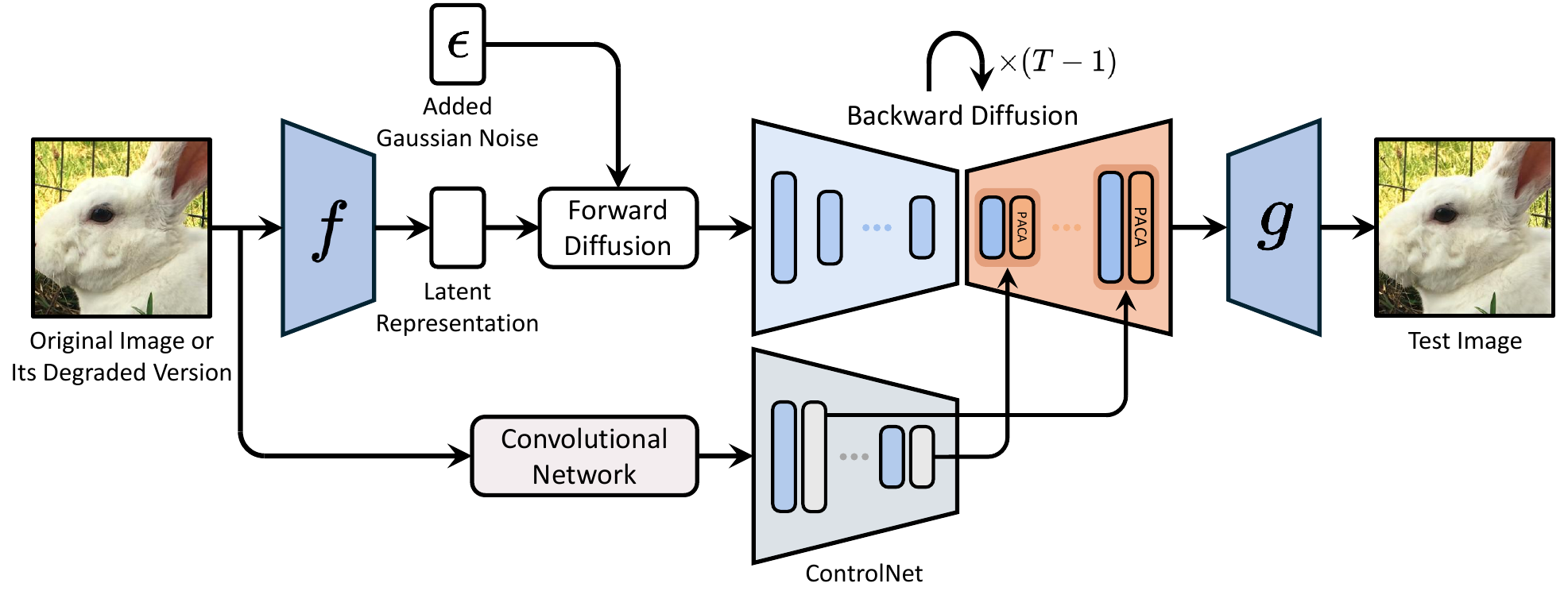}
    \caption{Inference of our generative image enhancer.}
	\label{fig:inference_arch}
\end{figure}

The overall architecture of our enhancer during inference is illustrated in Fig.~\ref{fig:inference_arch}.
As described in Sec.~\ref{sec:Construction of DiffIQA} of the main paper, 
we randomly applied the same degradations as used during training, augmented the initial image latent with additive Gaussian noise of varying intensities, and adjusted the sampling steps within range $[20, 1000]$.

Finally, we generated a total of $179,208$ test images using $20$ NVIDIA V100 GPUs, with an inference batch size of one per GPU. To accelerate inference, we employed the same UniPC Scheduler in PASD~\cite{yang2023pixel}. The entire inference process took approximately $20$ days.

\section{Subjective Experimental Setups of DiffIQA}
\label{training session for DiffIQA}
\begin{figure*}[h]
	\centering
	\includegraphics[width=0.8\linewidth]{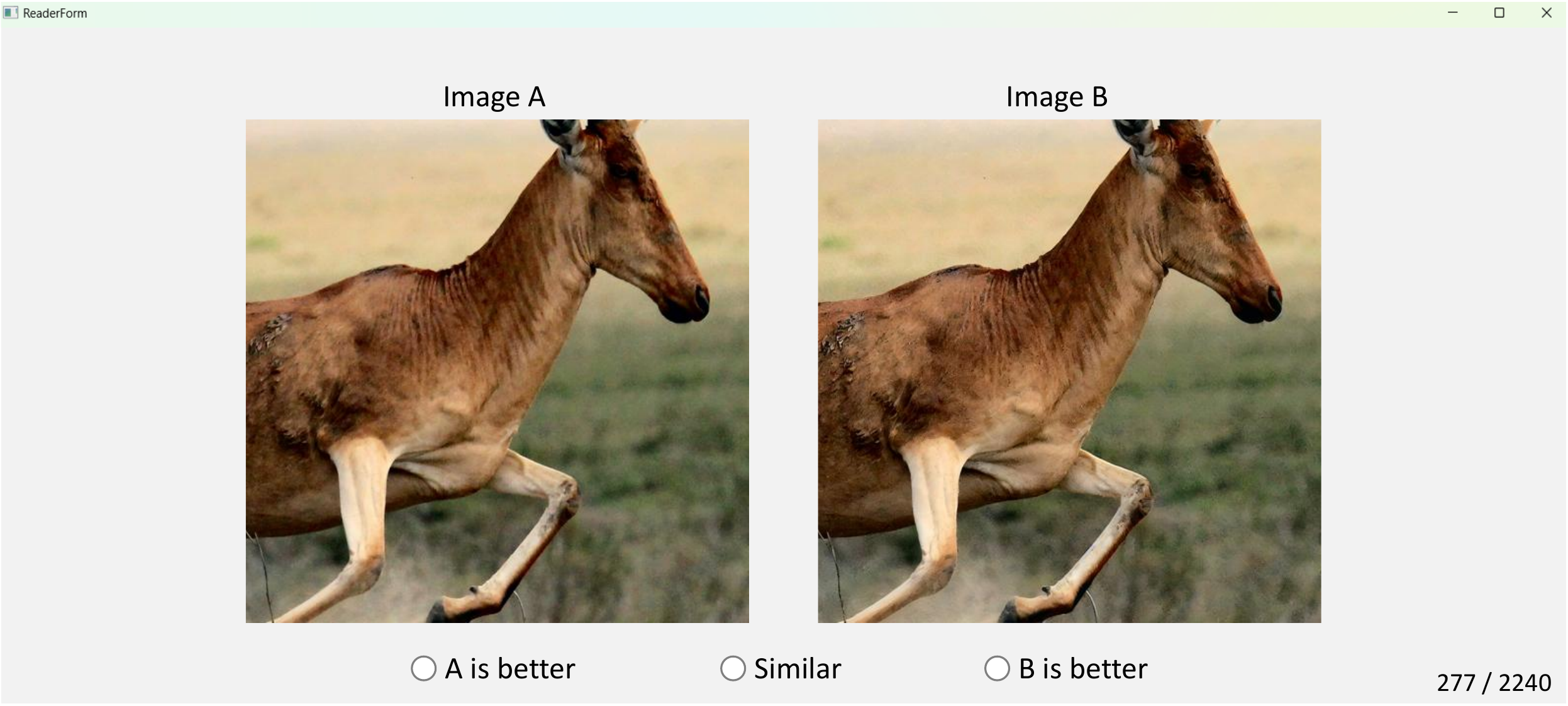}
    \caption{The GUI used for constructing DiffIQA.}
	\label{fig:software diffiqa}
\end{figure*}

We developed a graphical user interface (GUI), as illustrated in Fig.~\ref{fig:software diffiqa}, for MOS collection. This software is built using PyQt5\footnote{https://www.riverbankcomputing.com/software/pyqt/}, which is compatible with Windows Operating Systems from versions $8$ to $11$, ensuring low latency and support for screen resolutions ranging from 1,080 to 2K. Core functionalities of our GUI include 1) presentation of images in random spatial order; 2) a zoom-in feature using the mouse scroll wheel for more-detailed comparison; 3) a maximum of $10$-second viewing time with the prompt of the message: ``Please make your choice.''; 4) a radio button group of three choices; and 5) a checkpointing feature, ensuring that the subject can stop at any time to minimize the fatigue effect, and the software will resume from the last image pair when reopened. It is important to note that our paired comparison is incomplete, as the test image is compared solely with its reference image.

\begin{figure*}[t]
    \centering
    \begin{subfigure}[b]{0.16\textwidth}
        \includegraphics[width=\textwidth]{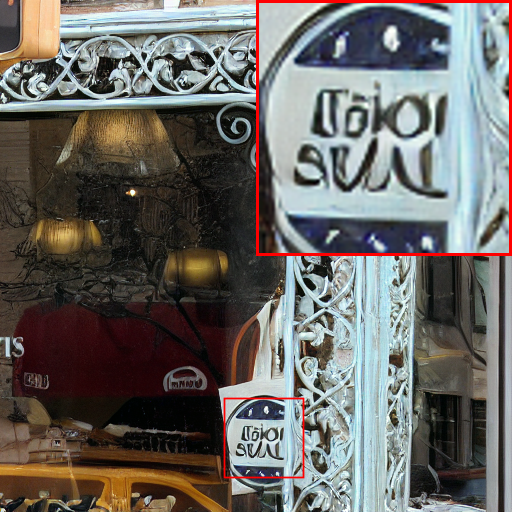}
        \caption{Generated}
    \end{subfigure}
    \begin{subfigure}[b]{0.16\textwidth}
        \includegraphics[width=\textwidth]{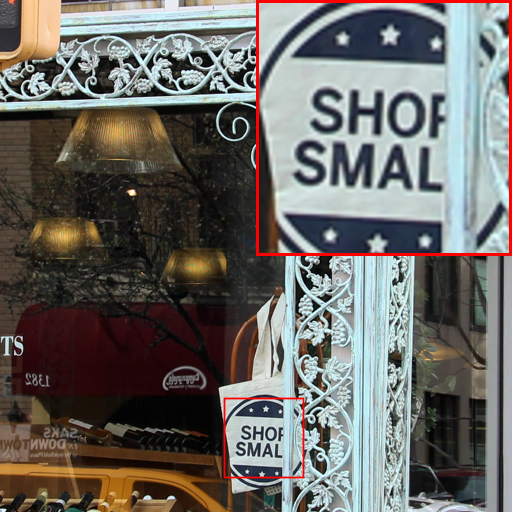}
        \caption{Reference}
    \end{subfigure}
    \begin{subfigure}[b]{0.16\textwidth}
        \includegraphics[width=\textwidth]{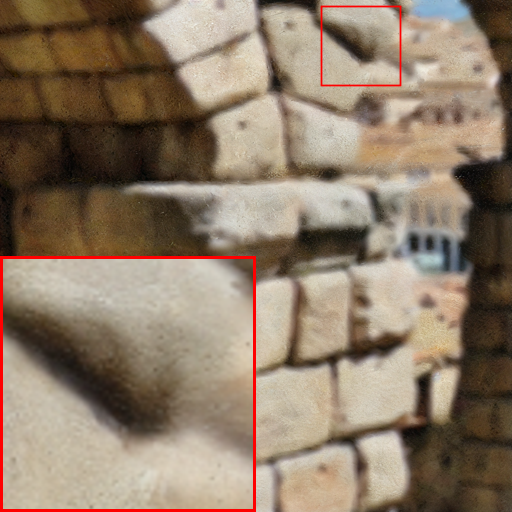}
        \caption{Generated}
    \end{subfigure}
    \begin{subfigure}[b]{0.16\textwidth}
        \includegraphics[width=\textwidth]{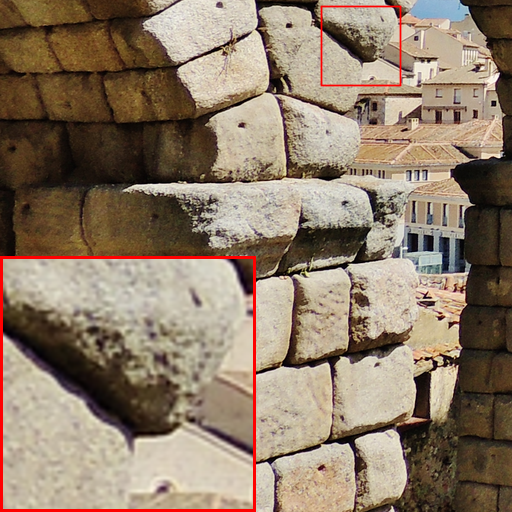}
        \caption{Reference}
    \end{subfigure}
    \begin{subfigure}[b]{0.16\textwidth}
        \includegraphics[width=\textwidth]{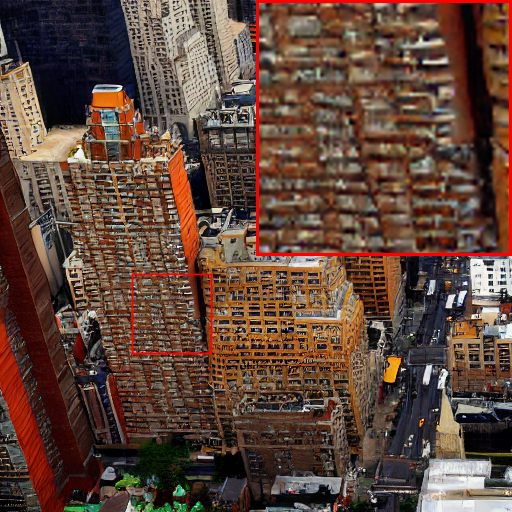}
        \caption{Generated}
    \end{subfigure}
    \begin{subfigure}[b]{0.16\textwidth}
        \includegraphics[width=\textwidth]{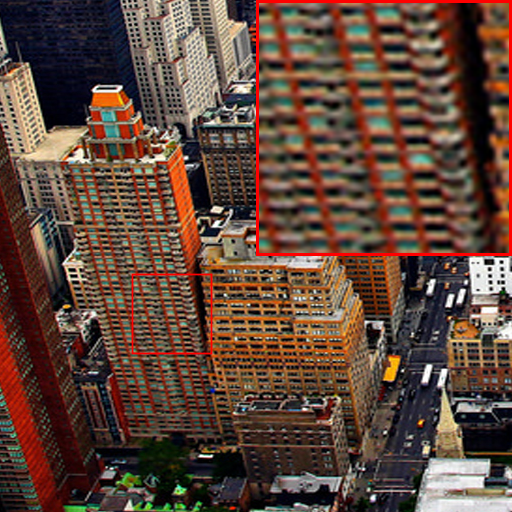}
        \caption{Reference}
    \end{subfigure}

    \vspace{0.2cm}

    \begin{subfigure}[b]{0.16\textwidth}
        \includegraphics[width=\textwidth]{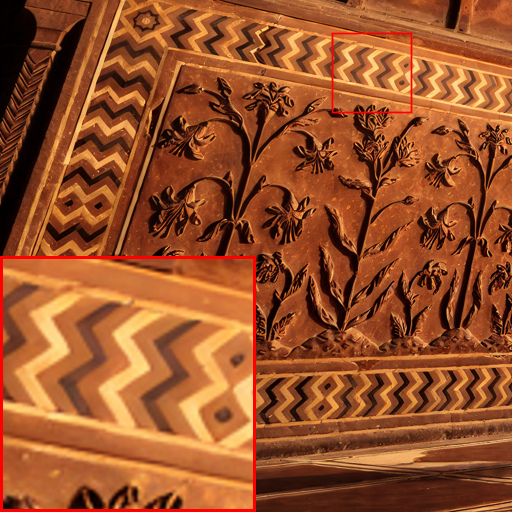}
        \caption{Generated}
    \end{subfigure}
    \begin{subfigure}[b]{0.16\textwidth}
        \includegraphics[width=\textwidth]{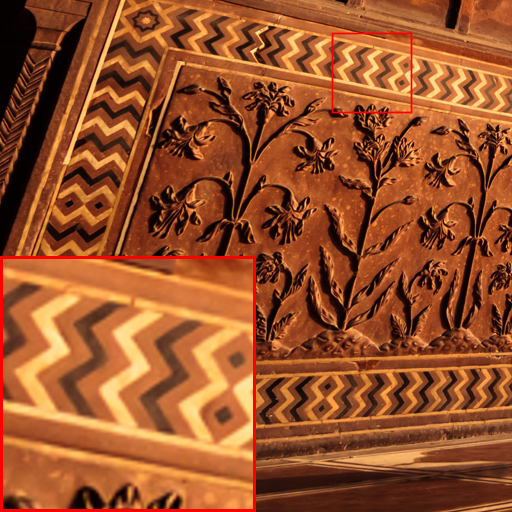}
        \caption{Reference}
    \end{subfigure}
    \begin{subfigure}[b]{0.16\textwidth}
        \includegraphics[width=\textwidth]{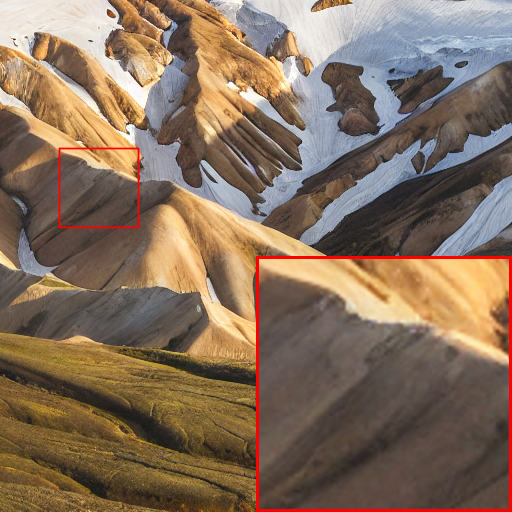}
        \caption{Generated}
    \end{subfigure}
    \begin{subfigure}[b]{0.16\textwidth}
        \includegraphics[width=\textwidth]{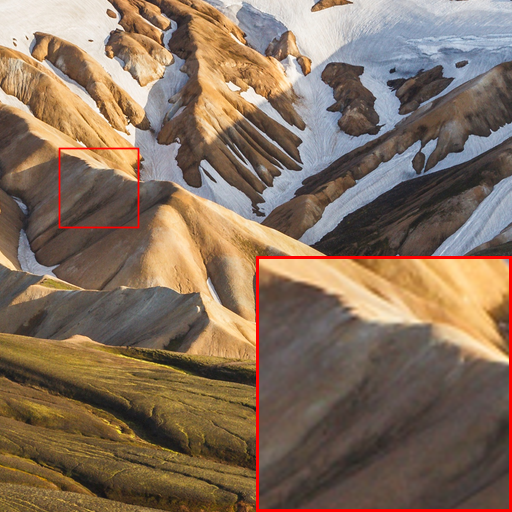}
        \caption{Reference}
    \end{subfigure}
    \begin{subfigure}[b]{0.16\textwidth}
        \includegraphics[width=\textwidth]{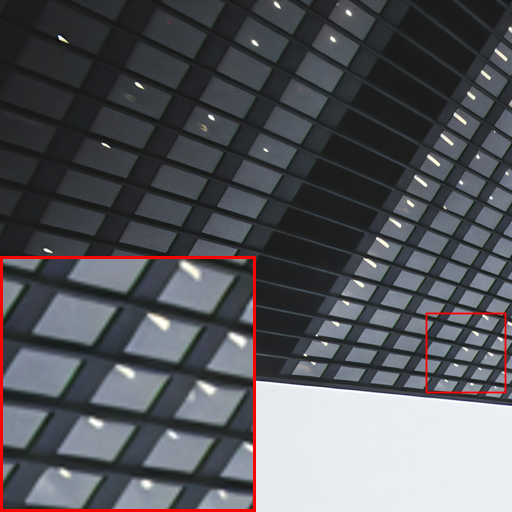}
        \caption{Generated}
    \end{subfigure}
    \begin{subfigure}[b]{0.16\textwidth}
        \includegraphics[width=\textwidth]{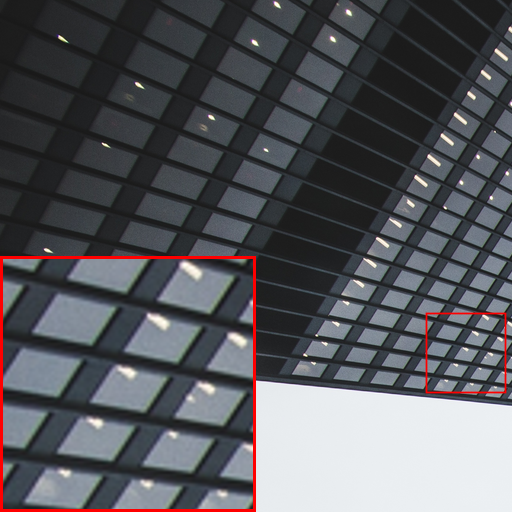}
        \caption{Reference}
    \end{subfigure}

    \vspace{0.2cm}

    \begin{subfigure}[b]{0.16\textwidth}
        \includegraphics[width=\textwidth]{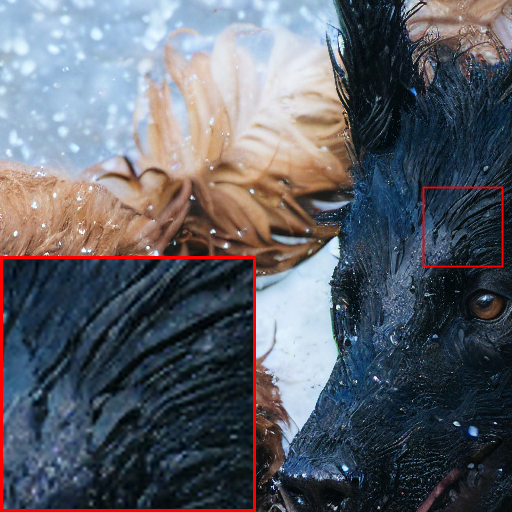}
        \caption{Generated}
    \end{subfigure}
    \begin{subfigure}[b]{0.16\textwidth}
        \includegraphics[width=\textwidth]{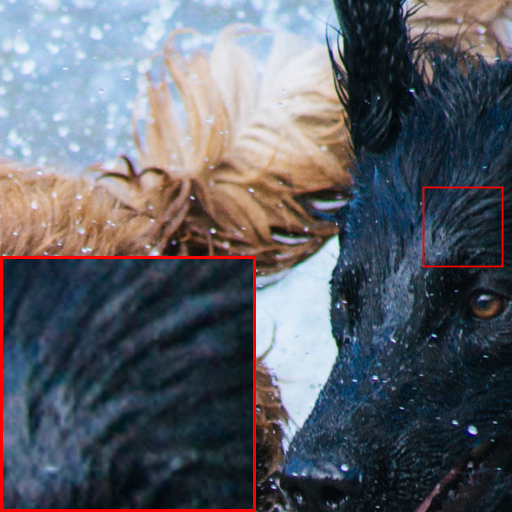}
        \caption{Reference}
    \end{subfigure}
    \begin{subfigure}[b]{0.16\textwidth}
        \includegraphics[width=\textwidth]{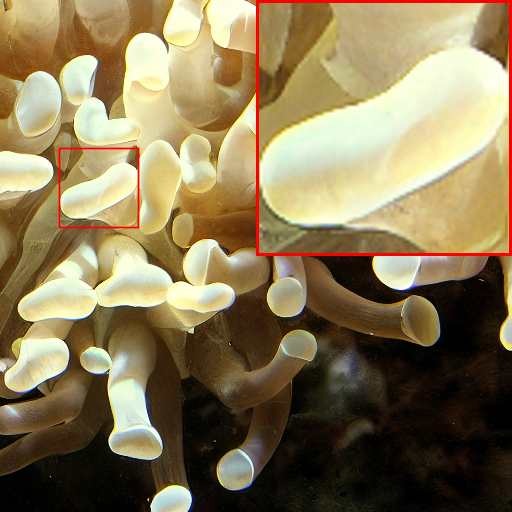}
        \caption{Generated}
    \end{subfigure}
    \begin{subfigure}[b]{0.16\textwidth}
        \includegraphics[width=\textwidth]{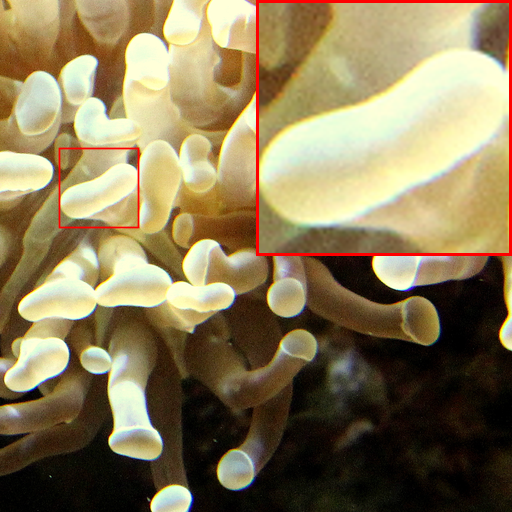}
        \caption{Reference}
    \end{subfigure}
    \begin{subfigure}[b]{0.16\textwidth}
        \includegraphics[width=\textwidth]{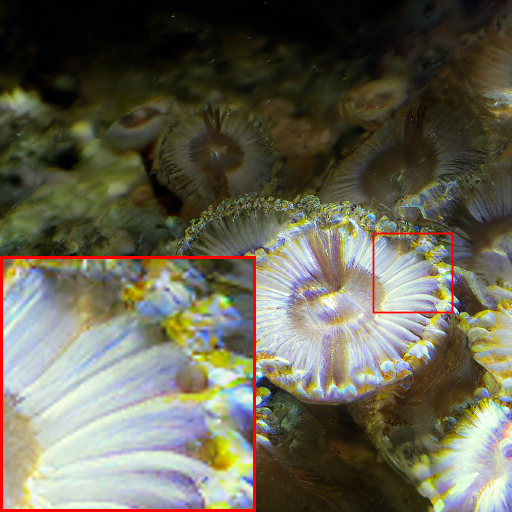}
        \caption{Generated}
    \end{subfigure}
    \begin{subfigure}[b]{0.16\textwidth}
        \includegraphics[width=\textwidth]{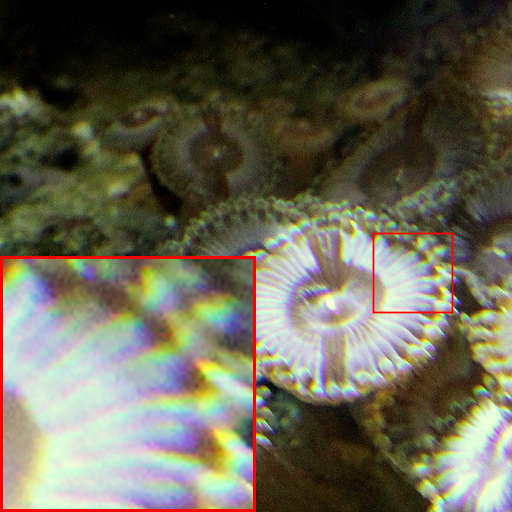}
        \caption{Reference}
    \end{subfigure}
    
    \caption{Representative images that are \textit{worse} (\textbf{(a)} to \textbf{(e)}), \textit{similar} (\textbf{(g)} to \textbf{(k)}), and \textit{better} (\textbf{(m)} to  \textbf{(q)}) relative to their references in our DiffIQA dataset.  Zoom in for better visibility.}
    \label{fig:diffiqa}
\end{figure*}

Before formal subjective testing, we included an approximately two-hour training session for all subjects, designed to familiarize them with the overall subjective testing procedure. Specifically, we provided a detailed demonstration of the specific functionalities of our GUI, and general guidelines to make visual comparisons. Subjects were instructed to focus primarily on image attributes closely related to perceived image quality,  such as image naturalness and distortion visibility, with some visual examples (see Fig.~\ref{fig:diffiqa}).

\begin{figure*}[h]
	\centering
	\includegraphics[width=0.8\linewidth]{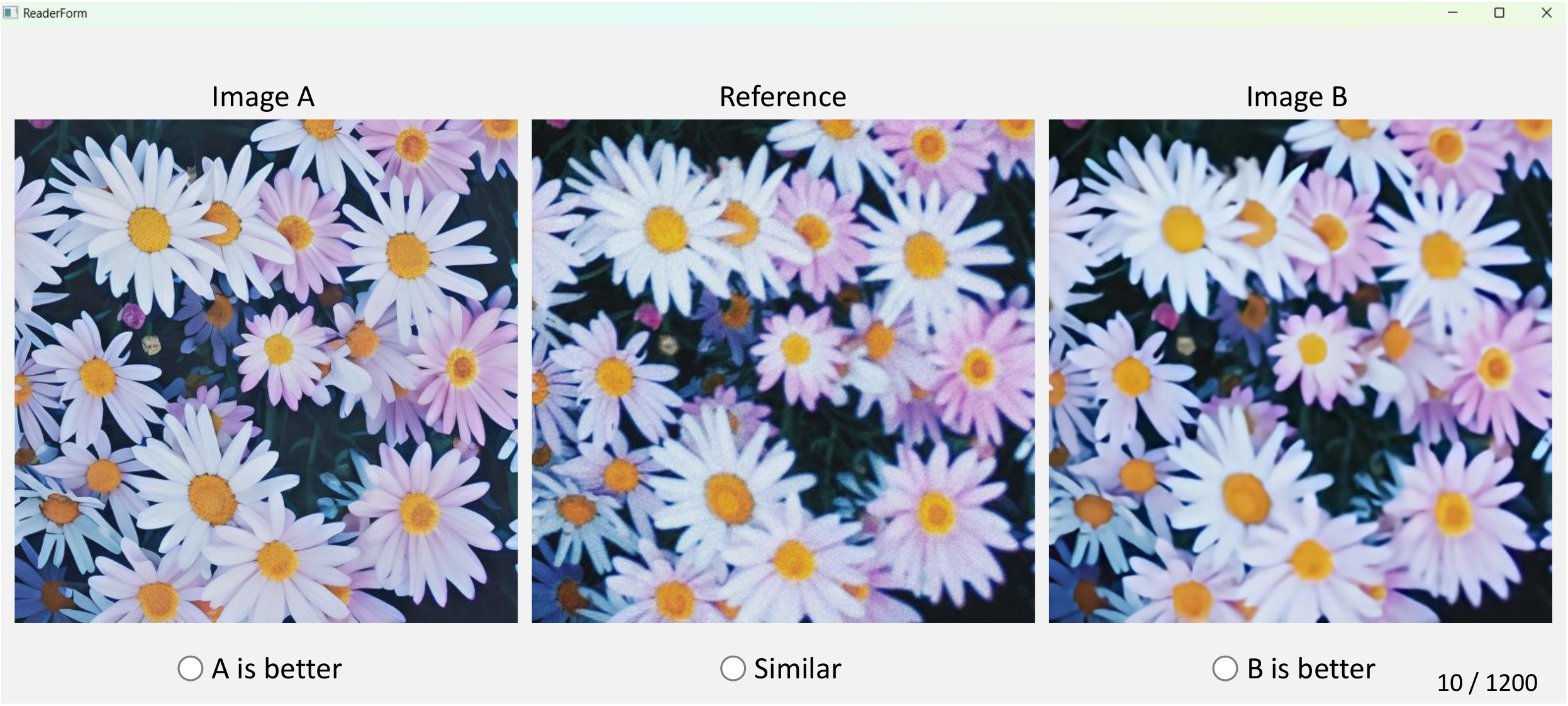}
    \caption{The GUI used for constructing SRIQA-Bench.}
	\label{fig:software sriqa}
\end{figure*}

\section{Subjective Experimental Setups of SRIQA-Bench}
The GUI for SRIQA-Bench closely resembles that of DiffIQA, with the key difference being the inclusion of a reference image in the middle for facilitating comparison of the two test images, as illustrated in Fig.~\ref{fig:software sriqa}.

Unlike the training session adopted in DiffIQA, subjects were first instructed to evaluate the fidelity of the two test images relative to the reference. If the test images exhibit comparable fidelity, subjects then selected the one with better quality, following similar guidelines described in Sec.~\ref{training session for DiffIQA}. Conversely, if the test images show significant differences in fidelity, subjects were instructed to choose the image with higher fidelity to the reference.

\section{Discussions on the Generated ``Fake''  Details}
\begin{figure*}[h]
  \centering
  \begin{subfigure}[b]{0.4\linewidth}
    \centering
    \includegraphics[width=\linewidth]{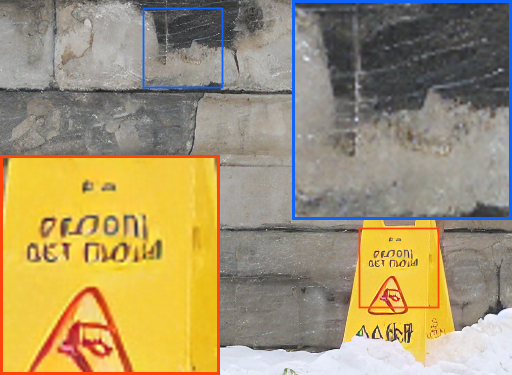}
    \caption{Test image}
  \end{subfigure}
  \begin{subfigure}[b]{0.4\linewidth}
    \centering
    \includegraphics[width=\linewidth]{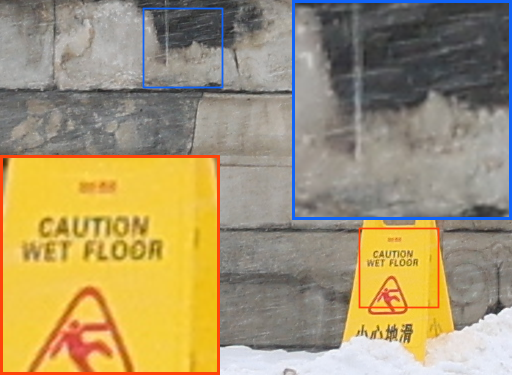}
    \caption{Reference image}
  \end{subfigure}
  \caption{Illustration of the ``fake'' generated details. In this example, the reference image is of better quality than the test image according to our subjective testing protocol.}
  \label{fig:distorted-details}
\end{figure*}

It is important to note that there are instances where the enhanced image appears to have superior overall quality, but the details differ significantly from the reference. This suggests that the enhanced details are hallucinated yet plausible. To address this issue during subjective testing, 
participants were instructed to prioritize deformed or fake details when assessing image quality. If such details impact the image’s fidelity, participants would annotate the image as having worse quality. As illustrated in Fig.~\ref{fig:distorted-details}, while the content in the blue box of the generated image appears sharper than the reference, the text in the red box is visibly distorted. Our model, A-FINE, correctly evaluates the reference image as having better quality, consistent with human judgments.

\end{document}

%% file: arxiv.bbl
\begin{thebibliography}{63}
\providecommand{\natexlab}[1]{#1}
\providecommand{\url}[1]{\texttt{#1}}
\expandafter\ifx\csname urlstyle\endcsname\relax
  \providecommand{\doi}[1]{doi: #1}\else
  \providecommand{\doi}{doi: \begingroup \urlstyle{rm}\Url}\fi

\bibitem[Chen et~al.(2024{\natexlab{a}})Chen, Mo, Hou, Wu, Liao, Sun, Yan, and
  Lin]{chen2024topiq}
Chaofeng Chen, Jiadi Mo, Jingwen Hou, Haoning Wu, Liang Liao, Wenxiu Sun, Qiong
  Yan, and Weisi Lin.
\newblock {TOPIQ}: A top-down approach from semantics to distortions for image
  quality assessment.
\newblock \emph{IEEE Transactions on Image Processing}, 33:\penalty0
  2404--2418, 2024{\natexlab{a}}.

\bibitem[Chen et~al.(2023)Chen, Liang, Zhang, Liu, Zeng, and
  Zhang]{chen2023human}
Du Chen, Jie Liang, Xindong Zhang, Ming Liu, Hui Zeng, and Lei Zhang.
\newblock Human guided ground-truth generation for realistic image
  super-resolution.
\newblock In \emph{IEEE/CVF Conference on Computer Vision and Pattern
  Recognition}, pages 14082--14091, 2023.

\bibitem[Chen et~al.(2024{\natexlab{b}})Chen, Zhang, Liang, and
  Zhang]{chen2024ssl}
Du Chen, Zhengqiang Zhang, Jie Liang, and Lei Zhang.
\newblock \uppercase{SSL}: A self-similarity loss for improving generative
  image super-resolution.
\newblock In \emph{ACM International Conference on Multimedia}, pages
  3189--3198, 2024{\natexlab{b}}.

\bibitem[Daly(1992)]{daly1992visible}
Scott~J Daly.
\newblock Visible differences predictor: An algorithm for the assessment of
  image fidelity.
\newblock In \emph{Human Vision, Visual Processing, and Digital Display III},
  pages 2--15, 1992.

\bibitem[Ding et~al.(2020)Ding, Ma, Wang, and Simoncelli]{ding2020image}
Keyan Ding, Kede Ma, Shiqi Wang, and Eero~P Simoncelli.
\newblock Image quality assessment: Unifying structure and texture similarity.
\newblock \emph{IEEE Transactions on Pattern Analysis and Machine
  Intelligence}, 44\penalty0 (5):\penalty0 2567--2581, 2020.

\bibitem[Ding et~al.(2021{\natexlab{a}})Ding, Liu, Zou, Wang, and
  Ma]{ding2021locally}
Keyan Ding, Yi Liu, Xueyi Zou, Shiqi Wang, and Kede Ma.
\newblock Locally adaptive structure and texture similarity for image quality
  assessment.
\newblock In \emph{ACM International Conference on Multimedia}, pages
  2483--2491, 2021{\natexlab{a}}.

\bibitem[Ding et~al.(2021{\natexlab{b}})Ding, Ma, Wang, and
  Simoncelli]{ding2021comparison}
Keyan Ding, Kede Ma, Shiqi Wang, and Eero~P Simoncelli.
\newblock Comparison of full-reference image quality models for optimization of
  image processing systems.
\newblock \emph{International Journal of Computer Vision}, 129\penalty0
  (4):\penalty0 1258--1281, 2021{\natexlab{b}}.

\bibitem[Dosovitskiy et~al.(2020)Dosovitskiy, Beyer, Kolesnikov, Weissenborn,
  Zhai, Unterthiner, Dehghani, Minderer, Heigold, Gelly, Uszkoreit, and
  Houlsby]{dosovitskiy2020image}
Alexey Dosovitskiy, Lucas Beyer, Alexander Kolesnikov, Dirk Weissenborn,
  Xiaohua Zhai, Thomas Unterthiner, Mostafa Dehghani, Matthias Minderer, Georg
  Heigold, Sylvain Gelly, Jakob Uszkoreit, and Neil. Houlsby.
\newblock An image is worth 16x16 words: Transformers for image recognition at
  scale.
\newblock \emph{arXiv preprint arXiv:2010.11929}, 2020.

\bibitem[Esser et~al.(2021)Esser, Rombach, and Ommer]{esser2021taming}
Patrick Esser, Robin Rombach, and Bjorn Ommer.
\newblock Taming {Transformers} for high-resolution image synthesis.
\newblock In \emph{IEEE/CVF Conference on Computer Vision and Pattern
  Recognition}, pages 12873--12883, 2021.

\bibitem[Gu et~al.(2020)Gu, Cai, Chen, Ye, Ren, and Dong]{gu2020pipal}
Jinjin Gu, Haoming Cai, Haoyu Chen, Xiaoxing Ye, Jimmy~S Ren, and Chao Dong.
\newblock {PIPAL}: A large-scale image quality assessment dataset for
  perceptual image restoration.
\newblock In \emph{European Conference on Computer Vision}, pages 633--651,
  2020.

\bibitem[Gu et~al.(2019)Gu, Lugmayr, Danelljan, Fritsche, Lamour, and
  Timofte]{gu2019div8k}
Shuhang Gu, Andreas Lugmayr, Martin Danelljan, Manuel Fritsche, Julien Lamour,
  and Radu Timofte.
\newblock {DIV8K}: Diverse {8K} resolution image dataset.
\newblock In \emph{IEEE/CVF International Conference on Computer Vision
  Workshop}, pages 3512--3516, 2019.

\bibitem[He et~al.(2016)He, Zhang, Ren, and Sun]{he2016deep}
Kaiming He, Xiangyu Zhang, Shaoqing Ren, and Jian Sun.
\newblock Deep residual learning for image recognition.
\newblock In \emph{IEEE/CVF Conference on Computer Vision and Pattern
  Recognition}, pages 770--778, 2016.

\bibitem[Karras et~al.(2019)Karras, Laine, and Aila]{karras2019style}
Tero Karras, Samuli Laine, and Timo Aila.
\newblock A style-based generator architecture for generative adversarial
  networks.
\newblock In \emph{IEEE/CVF Conference on Computer Vision and Pattern
  Recognition}, pages 4401--4410, 2019.

\bibitem[Lao et~al.(2022)Lao, Gong, Shi, Yang, Wu, Wang, Xia, and
  Yang]{lao2022attentions}
Shanshan Lao, Yuan Gong, Shuwei Shi, Sidi Yang, Tianhe Wu, Jiahao Wang, Weihao
  Xia, and Yujiu Yang.
\newblock Attentions help {CNNs} see better: Attention-based hybrid image
  quality assessment network.
\newblock In \emph{IEEE/CVF Conference on Computer Vision and Pattern
  Recognition Workshops}, pages 1140--1149, 2022.

\bibitem[Larson and Chandler(2010)]{larson2010most}
Eric~C Larson and Damon~M Chandler.
\newblock Most apparent distortion: Full-reference image quality assessment and
  the role of strategy.
\newblock \emph{Journal of Electronic Imaging}, 19\penalty0 (1):\penalty0
  1--21, 2010.

\bibitem[Ledig et~al.(2017)Ledig, Theis, Husz{\'a}r, Caballero, Cunningham,
  Acosta, Aitken, Tejani, Totz, Wang, and Shi]{ledig2017photo}
Christian Ledig, Lucas Theis, Ferenc Husz{\'a}r, Jose Caballero, Andrew
  Cunningham, Alejandro Acosta, Andrew Aitken, Alykhan Tejani, Johannes Totz,
  Zehan Wang, and Wenzhe Shi.
\newblock Photo-realistic single image super-resolution using a generative
  adversarial network.
\newblock In \emph{IEEE/CVF Conference on Computer Vision and Pattern
  Recognition}, pages 4681--4690, 2017.

\bibitem[Li et~al.(2023)Li, Zhang, Liang, Cao, Liu, Gong, Zhang, Tang, Liu,
  Demandolx, Ranjan, Timofte, and Van~Gool]{li2023lsdir}
Yawei Li, Kai Zhang, Jingyun Liang, Jiezhang Cao, Ce Liu, Rui Gong, Yulun
  Zhang, Hao Tang, Yun Liu, Denis Demandolx, Rakesh Ranjan, Radu Timofte, and
  Luc. Van~Gool.
\newblock \uppercase{LSDIR}: A large scale dataset for image restoration.
\newblock In \emph{IEEE/CVF Conference on Computer Vision and Pattern
  Recognition}, pages 1775--1787, 2023.

\bibitem[Liang et~al.(2021)Liang, Cao, Sun, Zhang, Van~Gool, and
  Timofte]{liang2021swinir}
Jingyun Liang, Jiezhang Cao, Guolei Sun, Kai Zhang, Luc Van~Gool, and Radu
  Timofte.
\newblock Swin\uppercase{IR}: Image restoration using \uppercase{S}win
  \uppercase{T}ransformer.
\newblock In \emph{IEEE/CVF International Conference on Computer Vision
  Workshop}, pages 1833--1844, 2021.

\bibitem[Liang et~al.(2022)Liang, Zeng, and Zhang]{liang2022details}
Jie Liang, Hui Zeng, and Lei Zhang.
\newblock Details or artifacts: A locally discriminative learning approach to
  realistic image super-resolution.
\newblock In \emph{IEEE/CVF Conference on Computer Vision and Pattern
  Recognition}, pages 5657--5666, 2022.

\bibitem[Lin et~al.(2019)Lin, Hosu, and Saupe]{lin2019kadid}
Hanhe Lin, Vlad Hosu, and Dietmar Saupe.
\newblock {KADID-10K}: A large-scale artificially distorted {IQA} database.
\newblock In \emph{IEEE International Conference on Quality of Multimedia
  Experience}, pages 1--3, 2019.

\bibitem[Loshchilov and Hutter(2017)]{loshchilov2017decoupled}
Ilya Loshchilov and Frank Hutter.
\newblock Decoupled weight decay regularization.
\newblock In \emph{International Conference on Learning Representations}, 2017.

\bibitem[Ma et~al.(2016)Ma, Duanmu, Wu, Wang, Yong, Li, and
  Zhang]{ma2016waterloo}
Kede Ma, Zhengfang Duanmu, Qingbo Wu, Zhou Wang, Hongwei Yong, Hongliang Li,
  and Lei Zhang.
\newblock Waterloo exploration database: New challenges for image quality
  assessment models.
\newblock \emph{IEEE Transactions on Image Processing}, 26\penalty0
  (2):\penalty0 1004--1016, 2016.

\bibitem[Mantiuk et~al.(2005)Mantiuk, Daly, Myszkowski, and
  Seidel]{mantiuk2005predicting}
Rafal~K Mantiuk, Scott~J Daly, Karol Myszkowski, and Hans-Peter Seidel.
\newblock Predicting visible differences in high dynamic range images: {Model}
  and its calibration.
\newblock In \emph{Human Vision and Electronic Imaging X}, pages 204--214,
  2005.

\bibitem[Mantiuk et~al.(2011)Mantiuk, Kim, Rempel, and
  Heidrich]{mantiuk2011hdr}
Rafal~K Mantiuk, Kil~Joong Kim, Allan~G Rempel, and Wolfgang Heidrich.
\newblock {HDR-VDP-2}: A calibrated visual metric for visibility and quality
  predictions in all luminance conditions.
\newblock \emph{ACM Transactions on Graphics}, 30\penalty0 (4):\penalty0 1--14,
  2011.

\bibitem[Mittal et~al.(2012)Mittal, Soundararajan, and Bovik]{mittal2012making}
Anish Mittal, Rajiv Soundararajan, and Alan~C Bovik.
\newblock Making a “completely blind” image quality analyzer.
\newblock \emph{IEEE Signal Processing Letters}, 20\penalty0 (3):\penalty0
  209--212, 2012.

\bibitem[Ponomarenko et~al.(2015)Ponomarenko, Jin, Ieremeiev, Lukin,
  Egiazarian, Astola, Vozel, Chehdi, Carli, Battisti, and
  C.-C]{ponomarenko2015image}
Nikolay Ponomarenko, Lina Jin, Oleg Ieremeiev, Vladimir Lukin, Karen
  Egiazarian, Jaakko Astola, Benoit Vozel, Kacem Chehdi, Marco Carli, Federica
  Battisti, and Kuo~J. C.-C.
\newblock Image database {TID2013}: Peculiarities, results and perspectives.
\newblock \emph{Signal Processing: Image Communication}, 30:\penalty0 57--77,
  2015.

\bibitem[Radford et~al.(2021)Radford, Kim, Hallacy, Ramesh, Goh, Agarwal,
  Sastry, Askell, Mishkin, Clark, Krueger, and Sutskever]{radford2021learning}
Alec Radford, Jong~Wook Kim, Chris Hallacy, Aditya Ramesh, Gabriel Goh,
  Sandhini Agarwal, Girish Sastry, Amanda Askell, Pamela Mishkin, Jack Clark,
  Gretchen Krueger, and Ilya Sutskever.
\newblock Learning transferable visual models from natural language
  supervision.
\newblock In \emph{International Conference on Machine Learning}, pages
  8748--8763, 2021.

\bibitem[Rombach et~al.(2022)Rombach, Blattmann, Lorenz, Esser, and
  Ommer]{rombach2022high}
Robin Rombach, Andreas Blattmann, Dominik Lorenz, Patrick Esser, and Bj{\"o}rn
  Ommer.
\newblock High-resolution image synthesis with latent diffusion models.
\newblock In \emph{IEEE/CVF Conference on Computer Vision and Pattern
  Recognition}, pages 10684--10695, 2022.

\bibitem[Sheikh and Bovik(2006)]{sheikh2006image}
Hamid~R Sheikh and Alan~C Bovik.
\newblock Image information and visual quality.
\newblock \emph{IEEE Transactions on Image Processing}, 15\penalty0
  (2):\penalty0 430--444, 2006.

\bibitem[Sheikh et~al.(2006)Sheikh, Sabir, and Bovik]{sheikh2006statistical}
Hamid~R Sheikh, Muhammad~F Sabir, and Alan~C Bovik.
\newblock A statistical evaluation of recent full reference image quality
  assessment algorithms.
\newblock \emph{IEEE Transactions on Image Processing}, 15\penalty0
  (11):\penalty0 3440--3451, 2006.

\bibitem[Simonyan and Zisserman(2015)]{simonyan2014very}
Karen Simonyan and Andrew Zisserman.
\newblock Very deep convolutional networks for large-scale image recognition.
\newblock In \emph{International Conference on Learning Representations}, 2015.

\bibitem[Sun et~al.(2023)Sun, Wu, Liang, Zhang, Yong, and
  Zhang]{sun2023improving}
Lingchen Sun, Rongyuan Wu, Jie Liang, Zhengqiang Zhang, Hongwei Yong, and Lei
  Zhang.
\newblock Improving the stability and efficiency of diffusion models for
  content consistent super-resolution.
\newblock \emph{arXiv preprint arXiv:2401.00877}, 2023.

\bibitem[Sun et~al.(2024)Sun, Wu, Ma, Liu, Yi, and Zhang]{sun2024pixel}
Lingchen Sun, Rongyuan Wu, Zhiyuan Ma, Shuaizheng Liu, Qiaosi Yi, and Lei
  Zhang.
\newblock Pixel-level and semantic-level adjustable super-resolution: A
  dual-{LoRA} approach.
\newblock \emph{arXiv preprint arXiv:2412.03017}, 2024.

\bibitem[Thurstone(1927)]{thurstone1927law}
Louis~L Thurstone.
\newblock A law of comparative judgment.
\newblock \emph{Psychological Review}, 34:\penalty0 273--286, 1927.

\bibitem[Tsai et~al.(2007)Tsai, Liu, Qin, Chen, and Ma]{tsai2007frank}
Ming-Feng Tsai, Tie-Yan Liu, Tao Qin, Hsin-Hsi Chen, and Wei-Ying Ma.
\newblock {FRank}: A ranking method with fidelity loss.
\newblock In \emph{International ACM SIGIR Conference on Research and
  Development in Information Retrieval}, pages 383--390, 2007.

\bibitem[Wang et~al.(2024{\natexlab{a}})Wang, Yue, Zhou, Chan, and
  Loy]{wang2024exploiting}
Jianyi Wang, Zongsheng Yue, Shangchen Zhou, Kelvin~CK Chan, and Chen~Change
  Loy.
\newblock Exploiting diffusion prior for real-world image super-resolution.
\newblock \emph{International Journal of Computer Vision}, pages 1--21,
  2024{\natexlab{a}}.

\bibitem[Wang et~al.(2015)Wang, Ma, Yeganeh, Wang, and Lin]{wang2015patch}
Shiqi Wang, Kede Ma, Hojatollah Yeganeh, Zhou Wang, and Weisi Lin.
\newblock A patch-structure representation method for quality assessment of
  contrast changed images.
\newblock \emph{IEEE Signal Processing Letters}, 22\penalty0 (12):\penalty0
  2387--2390, 2015.

\bibitem[Wang et~al.(2018{\natexlab{a}})Wang, Yu, Dong, and
  Loy]{wang2018recovering}
Xintao Wang, Ke Yu, Chao Dong, and Chen~Change Loy.
\newblock Recovering realistic texture in image super-resolution by deep
  spatial feature transform.
\newblock In \emph{IEEE/CVF Conference on Computer Vision and Pattern
  Recognition}, pages 606--615, 2018{\natexlab{a}}.

\bibitem[Wang et~al.(2018{\natexlab{b}})Wang, Yu, Wu, Gu, Liu, Dong, Qiao, and
  Loy]{wang2018esrgan}
Xintao Wang, Ke Yu, Shixiang Wu, Jinjin Gu, Yihao Liu, Chao Dong, Yu Qiao, and
  Chen~Change Loy.
\newblock {ESRGAN}: Enhanced super-resolution generative adversarial networks.
\newblock In \emph{European Conference on Computer Vision Workshops}, pages
  1--16, 2018{\natexlab{b}}.

\bibitem[Wang et~al.(2021)Wang, Xie, Dong, and Shan]{wang2021real}
Xintao Wang, Liangbin Xie, Chao Dong, and Ying Shan.
\newblock {Real-ESRGAN}: Training real-world blind super-resolution with pure
  synthetic data.
\newblock In \emph{IEEE/CVF International Conference on Computer Vision
  Workshops}, pages 1905--1914, 2021.

\bibitem[Wang et~al.(2024{\natexlab{b}})Wang, Yang, Chen, Wang, Guo, Chau, Liu,
  Qiao, Kot, and Wen]{wang2024sinsr}
Yufei Wang, Wenhan Yang, Xinyuan Chen, Yaohui Wang, Lanqing Guo, Lap-Pui Chau,
  Ziwei Liu, Yu Qiao, Alex~C Kot, and Bihan Wen.
\newblock {SinSR}: Diffusion-based image super-resolution in a single step.
\newblock In \emph{IEEE/CVF Conference on Computer Vision and Pattern
  Recognition}, pages 25796--25805, 2024{\natexlab{b}}.

\bibitem[Wang and Bovik(2009)]{wang2009mean}
Zhou Wang and Alan~C Bovik.
\newblock Mean squared error: Love it or leave it? {A} new look at signal
  fidelity measures.
\newblock \emph{IEEE Signal Processing Magazine}, 26\penalty0 (1):\penalty0
  98--117, 2009.

\bibitem[Wang and Bovik(2011)]{wang2011reduced}
Zhou Wang and Alan~C Bovik.
\newblock Reduced-and no-reference image quality assessment.
\newblock \emph{IEEE Signal Processing Magazine}, 28\penalty0 (6):\penalty0
  29--40, 2011.

\bibitem[Wang et~al.(2003)Wang, Simoncelli, and Bovik]{wang2003multiscale}
Zhou Wang, Eero~P Simoncelli, and Alan~C Bovik.
\newblock Multiscale structural similarity for image quality assessment.
\newblock In \emph{Asilomar Conference on Signals, Systems and Computers},
  pages 1398--1402, 2003.

\bibitem[Wang et~al.(2004)Wang, Bovik, Sheikh, and Simoncelli]{wang2004image}
Zhou Wang, Alan~C Bovik, Hamid~R Sheikh, and Eero~P Simoncelli.
\newblock Image quality assessment: From error visibility to structural
  similarity.
\newblock \emph{IEEE Transactions on Image Processing}, 13\penalty0
  (4):\penalty0 600--612, 2004.

\bibitem[Wu et~al.(2024{\natexlab{a}})Wu, Sun, Ma, and Zhang]{wu2024one}
Rongyuan Wu, Lingchen Sun, Zhiyuan Ma, and Lei Zhang.
\newblock One-step effective diffusion network for real-world image
  super-resolution.
\newblock \emph{arXiv preprint arXiv:2406.08177}, 2024{\natexlab{a}}.

\bibitem[Wu et~al.(2024{\natexlab{b}})Wu, Yang, Sun, Zhang, Li, and
  Zhang]{wu2024seesr}
Rongyuan Wu, Tao Yang, Lingchen Sun, Zhengqiang Zhang, Shuai Li, and Lei Zhang.
\newblock {SeeSR}: Towards semantics-aware real-world image super-resolution.
\newblock In \emph{IEEE/CVF conference on Computer Vision and Pattern
  Recognition}, pages 25456--25467, 2024{\natexlab{b}}.

\bibitem[Yang et~al.(2022)Yang, Wu, Shi, Lao, Gong, Cao, Wang, and
  Yang]{yang2022maniqa}
Sidi Yang, Tianhe Wu, Shuwei Shi, Shanshan Lao, Yuan Gong, Mingdeng Cao, Jiahao
  Wang, and Yujiu Yang.
\newblock {MANIQA}: Multi-dimension attention network for no-reference image
  quality assessment.
\newblock In \emph{IEEE/CVF Conference on Computer Vision and Pattern
  Recognition Workshops}, pages 1191--1200, 2022.

\bibitem[Yang et~al.(2024)Yang, Wu, Ren, Xie, and Zhang]{yang2023pixel}
Tao Yang, Rongyuan Wu, Peiran Ren, Xuansong Xie, and Lei Zhang.
\newblock Pixel-aware stable diffusion for realistic image super-resolution and
  personalized stylization.
\newblock In \emph{European Conference on Computer Vision}, pages 74--91, 2024.

\bibitem[Yeganeh et~al.(2015)Yeganeh, Rostami, and Wang]{yeganeh2015objective}
Hojatollah Yeganeh, Mohammad Rostami, and Zhou Wang.
\newblock Objective quality assessment of interpolated natural images.
\newblock \emph{IEEE Transactions on Image Processing}, 24\penalty0
  (11):\penalty0 4651--4663, 2015.

\bibitem[You et~al.(2025)You, Cai, Gu, Xue, and Dong]{you2025teaching}
Zhiyuan You, Xin Cai, Jinjin Gu, Tianfan Xue, and Chao Dong.
\newblock Teaching large language models to regress accurate image quality
  scores using score distribution.
\newblock \emph{arXiv preprint arXiv:2501.11561}, 2025.

\bibitem[Yu et~al.(2024)Yu, Gu, Li, Hu, Kong, Wang, He, Qiao, and
  Dong]{yu2024scaling}
Fanghua Yu, Jinjin Gu, Zheyuan Li, Jinfan Hu, Xiangtao Kong, Xintao Wang,
  Jingwen He, Yu Qiao, and Chao Dong.
\newblock Scaling up to excellence: Practicing model scaling for
  photo-realistic image restoration in the wild.
\newblock In \emph{IEEE/CVF Conference on Computer Vision and Pattern
  Recognition}, pages 25669--25680, 2024.

\bibitem[Zhang et~al.(2018{\natexlab{a}})Zhang, Cheng, and
  Hirakawa]{zhang2018corrupted}
Chen Zhang, Wu Cheng, and Keigo Hirakawa.
\newblock Corrupted reference image quality assessment of denoised images.
\newblock \emph{IEEE Transactions on Image Processing}, 28\penalty0
  (4):\penalty0 1732--1747, 2018{\natexlab{a}}.

\bibitem[Zhang et~al.(2021{\natexlab{a}})Zhang, Liang, Van~Gool, and
  Timofte]{zhang2021designing}
Kai Zhang, Jingyun Liang, Luc Van~Gool, and Radu Timofte.
\newblock Designing a practical degradation model for deep blind image
  super-resolution.
\newblock In \emph{IEEE/CVF International Conference on Computer Vision}, pages
  4791--4800, 2021{\natexlab{a}}.

\bibitem[Zhang et~al.(2011)Zhang, Zhang, Mou, and Zhang]{zhang2011fsim}
Lin Zhang, Lei Zhang, Xuanqin Mou, and David Zhang.
\newblock {FSIM}: A feature similarity index for image quality assessment.
\newblock \emph{IEEE Transactions on Image Processing}, 20\penalty0
  (8):\penalty0 2378--2386, 2011.

\bibitem[Zhang et~al.(2014)Zhang, Shen, and Li]{zhang2014vsi}
Lin Zhang, Ying Shen, and Hongyu Li.
\newblock {VSI}: A visual saliency-induced index for perceptual image quality
  assessment.
\newblock \emph{IEEE Transactions on Image processing}, 23\penalty0
  (10):\penalty0 4270--4281, 2014.

\bibitem[Zhang et~al.(2023{\natexlab{a}})Zhang, Rao, and
  Agrawala]{zhang2023adding}
Lvmin Zhang, Anyi Rao, and Maneesh Agrawala.
\newblock Adding conditional control to text-to-image diffusion models.
\newblock In \emph{IEEE/CVF International Conference on Computer Vision}, pages
  3836--3847, 2023{\natexlab{a}}.

\bibitem[Zhang et~al.(2018{\natexlab{b}})Zhang, Isola, Efros, Shechtman, and
  Wang]{zhang2018unreasonable}
Richard Zhang, Phillip Isola, Alexei~A Efros, Eli Shechtman, and Oliver Wang.
\newblock The unreasonable effectiveness of deep features as a perceptual
  metric.
\newblock In \emph{IEEE/CVF Conference on Computer Vision and Pattern
  Recognition}, pages 586--595, 2018{\natexlab{b}}.

\bibitem[Zhang et~al.(2019)Zhang, Liu, Dong, and Qiao]{zhang2019ranksrgan}
Wenlong Zhang, Yihao Liu, Chao Dong, and Yu Qiao.
\newblock {RankSRGAN}: Generative adversarial networks with ranker for image
  super-resolution.
\newblock In \emph{IEEE/CVF International Conference on Computer Vision}, pages
  3096--3105, 2019.

\bibitem[Zhang et~al.(2021{\natexlab{b}})Zhang, Ma, Zhai, and
  Yang]{zhang2021uncertainty}
Weixia Zhang, Kede Ma, Guangtao Zhai, and Xiaokang Yang.
\newblock Uncertainty-aware blind image quality assessment in the laboratory
  and wild.
\newblock \emph{IEEE Transactions on Image Processing}, 30:\penalty0
  3474--3486, 2021{\natexlab{b}}.

\bibitem[Zhang et~al.(2023{\natexlab{b}})Zhang, Zhai, Wei, Yang, and
  Ma]{zhang2023blind}
Weixia Zhang, Guangtao Zhai, Ying Wei, Xiaokang Yang, and Kede Ma.
\newblock Blind image quality assessment via vision-language correspondence: A
  multitask learning perspective.
\newblock In \emph{IEEE/CVF Conference on Computer Vision and Pattern
  Recognition}, pages 14071--14081, 2023{\natexlab{b}}.

\bibitem[Zheng et~al.(2021)Zheng, Yang, Fu, Zha, and Luo]{zheng2021learning}
Heliang Zheng, Huan Yang, Jianlong Fu, Zheng-Jun Zha, and Jiebo Luo.
\newblock Learning conditional knowledge distillation for degraded-reference
  image quality assessment.
\newblock In \emph{IEEE/CVF International Conference on Computer Vision}, pages
  10242--10251, 2021.

\bibitem[Zhou and Wang(2022)]{zhou2022quality}
Wei Zhou and Zhou Wang.
\newblock Quality assessment of image super-resolution: Balancing deterministic
  and statistical fidelity.
\newblock In \emph{ACM International Conference on Multimedia}, pages 934--942,
  2022.

\end{thebibliography}
